\documentclass[conference]{IEEEtran}

\def\BibTeX{{\rm B\kern-.05em{\sc i\kern-.025em b}\kern-.08em
    T\kern-.1667em\lower.7ex\hbox{E}\kern-.125emX}}

\usepackage{pifont}
\newcommand{\xmark}{\ding{55}}
\usepackage{booktabs}
\usepackage{mathtools}
\usepackage{braket}
\usepackage{multirow} 
\usepackage{enumitem}
\usepackage{makecell}
\usepackage{bigdelim, booktabs}

\usepackage{amsmath, amssymb} 
\usepackage{caption}          

\usepackage{inconsolata}
\usepackage[dvipsnames,table]{xcolor}
\colorlet{themecolor}{MidnightBlue!90!black}
\colorlet{highlight}{Orchid!90!black}
\usepackage[colorlinks,linkcolor=themecolor,citecolor=themecolor,urlcolor=themecolor,pagebackref=true]{hyperref}

\usepackage{orcidlink}

\begin{document}


\title{\textit{MALOQ}: \textit{M}assively \textit{A}ccelerated \textit{L}earning of \textit{O}perators for \textit{Q}uantum Transport}

\author{
\IEEEauthorblockN{
Manasa Kaniselvan\,\orcidlink{0000-0002-5331-8878},
Alexander Maeder\,\orcidlink{0009-0003-4420-5593},
Denghui Lu\,\orcidlink{0000-0003-0977-3635},
Alexandros Nikolaos Ziogas\,\orcidlink{0000-0002-4328-9751},
Mathieu Luisier\,\orcidlink{0000-0002-2212-7972}
}
\IEEEauthorblockA{\textit{D-ITET, ETH Zurich}, Zurich, Switzerland \\
\{mkaniselvan, almaeder, denghuilu, alziogas, mluisier\}@iis.ee.ethz.ch}
}







  
\maketitle

\begin{abstract}
Machine-learned (ML) operator models can be trained to predict density functional theory (DFT) Hamiltonian/density matrices at significantly reduced computational cost, thus extending electronic-structure calculations to previously unfeasible scales. Here, we introduce \textit{MALOQ} (Massively Accelerated Learning of Operators for Quantum Transport), an application built to train on and predict electronic-structure matrices for systems made of few to 100k atoms, described by large basis sets, and covering a wide range of atomic elements. Based on a state-of-the-art, SO(2)-equivariant backbone architecture, MALOQ provides (i) custom data-processing kernels to handle high-rank Hamiltonian matrix data and (ii) a scalable edge-wise distribution of atomic graph(s). Trained on the largest molecular Hamiltonian datasets available today, it reduces time-per-epoch by over 30\% compared to a molecule-wise-distributed framework, and enables inference on material graphs of arbitrary size. We demonstrate scalable training and inference for 3,000-12,000 atoms on the Alps supercomputer, up to 192 GPUs and 256 GPUs, respectively.

\end{abstract}


\section{Introduction}

Machine-learned (ML) models are accelerating computational materials science research by extending the reach of atomistic simulations to previously infeasible scales \cite{Yuan2026}. So far, these models have focused on learning \textit{properties} at the molecular and atomic level. This includes the class of ML Interatomic Potentials (MLIPs), which predict the molecular energies and atomic forces used in molecular dynamics (MD) simulations. State-of-the-art, graph-based `materials foundation models' are now capable of generating these quantities for arbitrary molecules and complex material geometries at quasi-DFT-level accuracy \cite{mace, uma, pet}. 

An emerging class of ML materials models extends these architectures toward electronic-level quantities, defined by \textit{operators} \cite{deeph3, helm, nigam2026, wanet}. Rather than learning atom-centered properties, these `operator models' are directly trained on the matrices encoding interactions between the inter- and intra-atomic orbitals of a given system. The most commonly targeted operators are the Hamiltonian/Fock ($\mathbf{H}$) and density ($\mathbf{P}$) matrices, which describe the spatial and energetic distribution of states that electrons can occupy within the atomic structure considered. Recent works have extended these principles to learn other operators that can be represented in an atomic orbital basis, either directly, such as the Green's function (inverse of $\mathbf{H}$) \cite{Venturella2025}, or indirectly, e.g., lower-rank Hessian \cite{hip} matrices which are derived from forces. Operator models are being developed for two main purposes:

\begin{figure*}[!t]
  \centering
  \includegraphics[width=\linewidth]{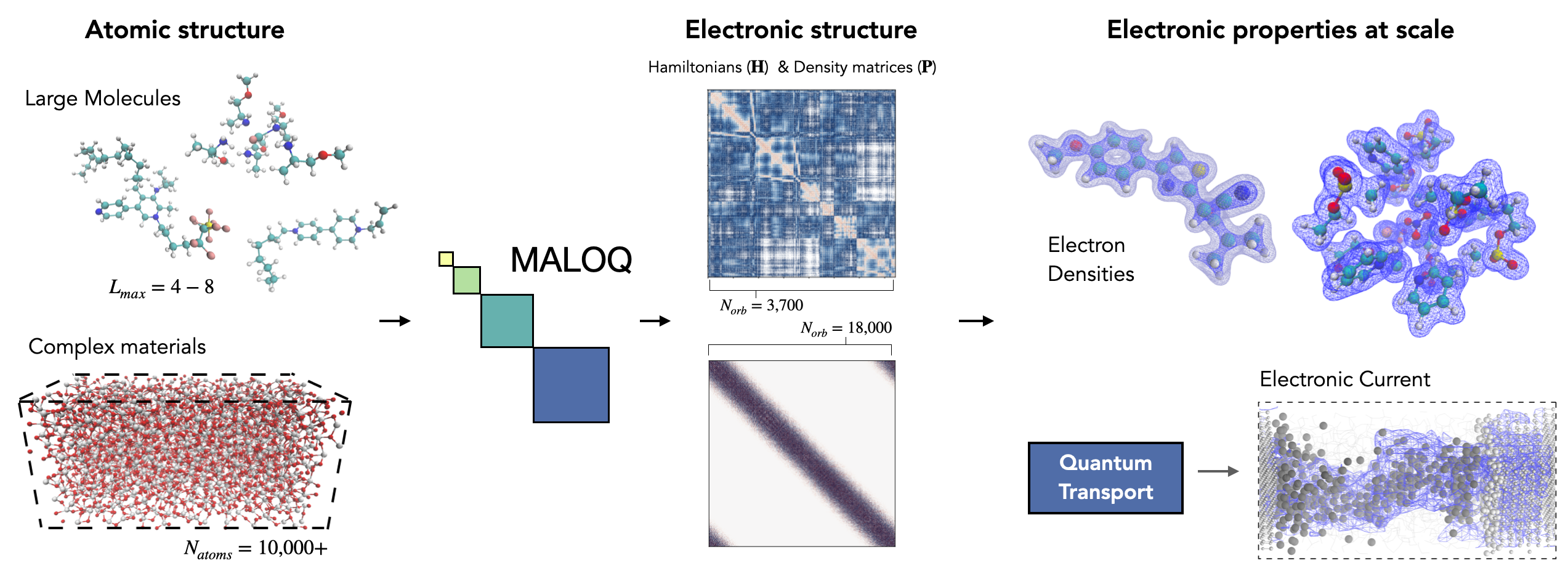}
  \caption{Applications of electronic structure learning in molecules (top) and materials (bottom). The goal is to learn the mapping between atomic structures (left) and their corresponding electronic Hamiltonian matrices (middle), and predict them at scales inaccessible by DFT alone. From these matrices, we can analytically construct electron densities or compute the electronic current flowing through complex materials (right).}
  \label{fig:matrices}
\end{figure*}

\begin{enumerate}[leftmargin=15pt, listparindent=0pt, parsep=0pt]

\vspace{0.1in}

\item \textbf{Enabling quantum transport through complex materials:} The Hamiltonian in matrix form serves as an input to quantum transport (QT) simulations that aim to compute the electronic current flowing through atomically-resolved devices under different voltage configurations. Advanced QT solvers operate on electronic structure inputs at the level of density functional theory (DFT), one of the most powerful \textit{ab initio} methods for electronic structure calculations \cite{Brandbyge2002}. Currently, QT codes are capable of treating systems with 10k+ atoms, in the presence of complex many-body effects \cite{Vetsch2025}, matching the physical dimensions of electronic devices fabricated by semiconductor companies \cite{intel-nrfet}. However, generating $\mathbf{H}$ using DFT codes scales with $\mathcal{O}(N_{orb}^3)$, $N_{orb}$ being the total number of atomic orbitals in the structure of interest, ultimately limiting the problem sizes (number of atoms) that can be handled. Existing works typically compute $\mathbf{H}$ for small, periodic systems and tile them to construct device-scale models. Meanwhile, these operator models can generate $\mathbf{H}$ with $\mathcal{O}(N_{orb})$ scalability \cite{Tang2024}. In the same way that MLIPs have enabled new regimes of MD simulations, operator models thus promise to unlock QT modeling of realistically complex atomic geometries \cite{amorphous}. 

\vspace{0.1in}

\item \textbf{Exploiting scaling laws for downstream properties:} ML property models are data-limited - their performance improves with more training data \cite{uma}. However, the forces and energies in existing training datasets have already consumed billions of core hours to be assembled at the DFT level \cite{omol}, and generating orders-of-magnitude more data is not presently feasible. Within these datasets, the processing of each molecule or material unit cell of $N$ atoms generates $\mathcal{O}(1)$ energy labels, $\mathcal{O}(N)$ force labels, and $\mathcal{O}(N^2)$ orbital interaction labels in the electronic structure operators. Leveraging the latter available but so-far-ignored electronic data through pre-training schemes provides an efficient route to continue exploiting `neural scaling laws,' which define the relationship between training dataset size and model accuracy. In DFT, the orbital interactions within electronic structure operators parameterize all subsequent atomic-level molecular and material properties. Incorporating this information into \textit{property} prediction architectures improves generalization on atomic- and molecular-prediction tasks \cite{hip, helm, orbitall, Suman2025, nigam2026}. Small molecule benchmarks have shown that computing forces and energies analytically from learned electronic structure operators leads to accuracy competitive with direct predictions using MLIPs \cite{Seongsu2026}.

\end{enumerate}

Sizable datasets of electronic structure matrices have recently become available, with up to 10 PB \cite{omol} of orbital interactions. Training on this matrix data, however, introduces a different set of computational challenges than previously encountered by ML property models such as MLIPs. First, the spherical features used within embeddings must match the dimensions of high-rank orbital interaction data, typically up to $f$ (and sometimes $g$) orbitals (requiring angular degree $l=6-8$). Contrary to MLIPs where the rank of the maximum spherical coefficients is typically a hyperparameter, electronic structure prediction requires high-rank embeddings simply to produce data of correct dimensions. Secondly, while MLIPs typically enforce cutoffs of $\sim$6-8 \AA\; between interacting atoms, treating orbital interaction data requires handling dedicated edge embeddings that extend to interatomic distances of 12 \AA\; \cite{amorphous, helm}. Larger cutoffs create densely connected graph representations with node degrees over 100, significantly increasing the total volume of embeddings to be processed. Finally, processing orbital interaction data necessitates frequent basis transformations during training and inference to convert between `label' and `matrix' representations of $\mathbf{H}$. Together, these three key differences substantially increase the memory footprint of both training and inference. 


To enable operator prediction `at scale', we therefore introduce the \textit{MALOQ} (\textit{M}assively \textit{A}ccelerated \textit{L}earning of \textit{O}perators for \textit{Q}uantum Transport) package, an accelerated ML framework to train electronic structure matrices and predict them for arbitrary atomic elements and structure sizes, from large molecules to 10k+ atom material systems (\textbf{Fig.~\ref{fig:matrices}}). At the core of MALOQ is a message passing Graph Neural Network (GNN) leveraging equivariant spherical channel network (eSCN) convolutions \cite{so2}. Its backbone architecture has achieved state-of-the-art performance for both MLIPs (eSEN \cite{eSEN}) and Hamiltonian prediction tasks (HELM \cite{helm}). On top of it, we introduce two categories of computational innovations designed to enable massively parallel training and inference of electronic structure matrices:

\begin{enumerate}
\item Operator data processing acceleration: \textbf{(1a)} Efficient `Matrix $\leftrightarrow$ Label' conversions kernels, allowing for Hamiltonian matrices to be pre-processed into training labels at $\sim10^6$ atoms/s, and rapidly reconstructed from predicted spherical coefficients. \textbf{(1b)} Accelerated computation of the generalized rotation matrices (Wigner-D matrices) required for eSCN convolutions.
\item Custom distribution of large molecular and material atomic graphs to accelerate training over molecular datasets, and perform inference for the $\mathbf{H}$ of large atomic systems. We combine an efficient point-to-point communication scheme with edge-wise partitioning to minimize communication and reduce load imbalance between partitions, achieving strong scaling up to 192 GH200 GPUs, and $>90\%$ weak scaling efficiency on structure sizes of 12,000 atoms ($\sim$10 million orbital interactions).
\end{enumerate}

Together, these contributions enable us to \textbf{(1)} improve training speeds over irregular molecular databases by 30-50\%, and \textbf{(2)} achieve rapid inference for Hamiltonian matrices on large material graphs containing 10k+ atoms. On the one hand, our work allows for the simulation of industry-relevant atomic systems whose quantum transport properties can now be readily determined at the DFT level with reduced computational complexity. On the other hand, it enables an efficient treatment of the tremendous volume of molecular and material property data available at electronic resolution, creating an avenue for ML property models to continue benefiting from neural scaling laws without additional dataset generation.

\section{Background \& Related Work}
\label{sec:background}

\subsection{The electronic structure problem}

The training data for MLIPs and other property models is typically generated using DFT. This first principle method implements the Kohn-Sham equations, which self-consistently couple the electron density and electrostatic potential of the atomic system of interest until convergence is reached, after $N_{iter}$ iterations \cite{Kohn1965}. Several commercially or freely available codes implement this self-consistent and iterative ``density$\leftrightarrow$potential'' scheme on atomic structures \cite{cp2k, orca}. 

Many of these implementations operate in a basis of localized atomic orbitals $|\varphi_i\rangle$, often constructed from contracted Gaussian functions \cite{cp2k, orca} that transform like spherical harmonics (\( Y^{l}_m(\hat{r}')\)) under rotation. Each spherical harmonic is specified by an angular momentum degree $l$ and order \( m \in \{-l, \dots, l\} \), e.g., $s (l=0)$, $p (l=1)$, $d (l=2)$, $\cdots$. In this basis, the Kohn-Sham equations take the form of a generalized eigenvalue problem: \(\textbf{H} \psi = \varepsilon \textbf{S} \psi\). The Hamiltonian matrix $\mathbf{H}$ has entries $H_{ij}=\langle \varphi_i | \hat{H}(\mathbf{r}) | \varphi_j \rangle$ where $\hat{H}(\mathbf{r})$ is the Hamiltonian operator. The overlap matrix $\mathbf{S}$ is made of terms $S_{ij}=\langle \varphi_i | \varphi_j \rangle $ that describe finite overlaps between the localized orbitals. They can be pre-computed from the orbital basis. Both $\mathbf{H}$ and $\mathbf{S}$ are matrices of size \(N_{orb} = \sum_{k=1}^{N_A} N_{orb,k} \), where \(N_{A}\) is the total number of atoms, \(N_{orb,k}\) the number of orbitals (basis elements) of atom $k$, and the index $k$ runs over all atoms accounted for. The solution of the Kohn-Sham equations provides the energy eigenvalues ($\varepsilon$) and wavefunctions ($\psi$) from which the charge density $\rho(\mathbf{r})$ can be derived. The core operation within DFT codes then consists of repeatedly solving the aforementioned generalized eigenvalue problem, until self-consistent convergence is reached between $\rho(\mathbf{r})$ and the resulting electrostatic potential $V(\mathbf{r})$. The computational complexity of the full method scales with \(\mathcal{O}(N_{orb}^3)\) per iteration. At the end of these self-consistent iterations, atomic- and molecular-level properties, such as forces and energies, can be computed from the final, converged $\mathbf{H}$. 


\subsection{Related work}

\setlength{\tabcolsep}{3pt} 
\begin{table}[!t]
\centering
\begin{tabular}{c c l l l c c c r} 
\toprule
& & \textbf{Application} & \textbf{Op.} & \textbf{Task} & \multicolumn{2}{c}{\textbf{Prediction}} & \textbf{Distr.} & $\mathbf{N_A}$ ($L_{max}$) \\
\cmidrule(lr){6-7}
& & & & & \textbf{Node} & \textbf{Edge} & & \\
\midrule

\parbox[t]{2mm}{\multirow{5}{*}{\rotatebox[origin=c]{90}{\scriptsize MLIPs}}} 
& \ldelim\{{5}{4pt} & DeepMD \cite{deepmd} & Linear & F/E & \checkmark & \xmark & \checkmark & - \\
& & Allegro \cite{Kozinsky2023} & TP & F/E & \checkmark & \xmark & \checkmark & - \\
& & MACE \cite{mace} & TP & F/E & \checkmark & \xmark & \xmark & - \\
& & SevenNet \cite{Park2024} & TP & F/E & \checkmark & \xmark & \checkmark & - \\
& & UMA \cite{uma} & \textbf{eSCN} & F/E & \checkmark & \xmark & \checkmark & - \\

\midrule

\parbox[t]{2mm}{\multirow{5}{*}{\rotatebox[origin=c]{90}{\scriptsize Hamiltonians}}} 
& \ldelim\{{5}{4pt} & QHNet \cite{quantumham} & TP & $H_{ij}$ & \checkmark & \checkmark & \xmark & 10 (4) \\
& & WANet \cite{wanet} & \textbf{eSCN} & $H_{ij}$ & \checkmark & \checkmark & \xmark & 100 (6) \\
& & HELM \cite{helm} & \textbf{eSCN} & $H_{ij}$ & \checkmark & \checkmark & \xmark & 150 (6) \\
& & DeepH \cite{deeph-pack} & \textbf{eSCN} & $H_{ij}$ & \checkmark & \checkmark & \xmark & 216 (4) \\
& & \textbf{MALOQ} & \textbf{eSCN} & $H_{ij}$ & \checkmark & \checkmark & \checkmark & \textbf{3,000+ (4-8)} \\

\bottomrule
\end{tabular}
\caption{Selection of GNN applications for materials modeling across atomic elements (near-`universal'), including functional similarities and differences in the prediction tasks, operation (Op.) used to update node/edge features, and possibility of distributed compute environment (Distr). F/E = Forces \& Energies, $H_{ij}$ = Hamiltonian, TP = SO(3) equivariant tensor product, eSCN = SO(2) equivariant spherical channel network. $\mathbf{N_A}$ = Maximum \textit{training} structure size.}
\label{tab:tab}
\end{table}

GNNs are now mainstream in computational materials science, where they are used to learn structure-property relationships on atomic graphs. Training occurs through message passing, where nodes and/or edges of the graph are updated as a learnable function of other nodes/edges (typically atomic elements/interatomic distance) from their neighborhoods. Successive message passing (MP) layers gradually transform the initial identity of each node/edge such that they can be mapped to targets representing material properties. 

The success of many GNNs in learning structure-property relationships hinges on preserving the structural symmetries of the input atomic graphs. Graph-based architectures, \textit{per se}, ensure permutation and translation invariance of atomic coordinates. Imposing interactions cutoffs allows models to further take advantage of the physical nearsightedness of atomic-interactions. Rotational symmetries have been incorporated by explicitly designing the GNNs to commute with rotation operations, typically by using tensor products and gated activations in place of linear layers and nonlinear activation functions. Compared to otherwise `rotationally invariant' networks, which depend on data-augmentation to learn these relationships, the `rotationally equivariant' models considered here learn physically-meaningful mappings with less data \cite{Batzner2022} and lower compute budgets \cite{brehmer2024doesequivariancematterscale}. While models without encoded rotational symmetries can perform competitively when trained on sufficient property data \cite{pet}, for operators such as electronic structure matrices, the data explicitly takes the form of high-rank embeddings, making incorporation of rotational symmetries unavoidable for sufficient accuracy \cite{helm, wanet}. 


We present key features of state-of-the-art GNNs for MLIP and Hamiltonian prediction in \textbf{Table~\ref{tab:tab}}, and compare them to MALOQ. Note that GNNs targeting MLIPs are restricted to `node' prediction tasks, while those dedicated to Hamiltonian matrices must also include `edge' prediction capabilities. For all models performing electronic structure prediction, we include the maximum system sizes they have so far been reported to achieve. Two aspects in \textbf{Table \ref{tab:tab}} are particularly relevant to our work, (1) the development of models specific to Hamiltonian prediction and (2) domain decomposition/distribution over the underlying atomic graphs. Below, we discuss their progression and current state of the art.

\begin{table*}[t]
\small
\centering
\label{tab:dataset_properties}
\renewcommand{\arraystretch}{1.2}
\setlength{\tabcolsep}{5pt} 
\begin{tabular}{l l l c c c c c c c}
\toprule
\textbf{} &
\textbf{Dataset} &
\textbf{Elements} &
\textbf{$\mathbf{N_{atom}^{total}}$} &
\textbf{$\mathbf{N_{atom}^{avg}}$} &
\textbf{$\mathbf{d_{node}^{avg}}$} & 
\textbf{Basis} ($\mathbf{L_{\max}}$) &
\textbf{Func.} &
\textbf{$r_{cut}$ (\AA)} \\
\midrule

\multirow{3}{*}{\makecell{Small \\ molecule}}
& MD17
& H, C, N, O
& -
& -
& (FC) 
& def2\_SVP (\textbf{4})
& PBE
& (FC)\\

& QH9
& H, C, N, O, F
& -
& -
& (FC) 
& def2\_SVP (\textbf{4})
& B3LYP
& (FC) \\

& \textbf{$\nabla^2$DFT (2k split)}
& H, C, N, O, F, S, Cl, Br
& 938,920
&  77
& (FC) 
& def2\_SVP (\textbf{4})
& $\omega$B97X-D
& (FC) \\

\midrule
\multirow{3}{*}{\makecell{Large \\ Molecule}}

& CSH\_58k
& $\times$58 elements
& - 
& -
& -
& def2-TZVPD (\textbf{6})
& $\omega$B97M-V
& 12.0 \\

& \textbf{Electrolytes}
& $\times$19 elements
& 1,201,369
& 41
& 48
& def2-TZVPD (\textbf{6})
& $\omega$B97M-V
& 12.0 \\

& \textbf{Metal-Organics}
& $\times$31 elements
& 1,064,625
& 75
& 79
& def2-TZVPD (\textbf{8})
& $\omega$B97M-V
& 12.0 \\

\midrule
\multirow{2}{*}{\makecell{Material}}
& \textbf{Amorphous-HfO$_2$}
& Hf, O
& 9,000
& 3,000
& 598
& SZV/DZVP (\textbf{4})
& PBE
& 12.0 \\

& Amorphous-GST
& Ge, Sb, Te
& -
& -
& - 
& DZVP (\textbf{4})
& PBE
& 12.0 \\

\bottomrule
\end{tabular}
\caption{Summary of existing public Hamiltonian matrix datasets, highlighting the distribution of structure size (\# atoms, $\mathbf{N_{atom}^{total}}$, $\mathbf{N_{atom}^{average}}$), node degree ($\mathbf{d_{node}}$), basis set and $\mathbf{L_{max}}$ required to describe it, functional type (Func.), and interaction cutoff radius. `FC' means `fully connected graph', in case molecules are small enough to avoid applying a cutoff. Data sources: MD17 \cite{schnorb}, QH9 \cite{qh9dataset}, $\nabla^2$DFT \cite{nablaDFT}, amorphous materials \cite{amorphous}, electronic structure subsets from OMol25 \cite{omol, helm}.}
\label{tab:dataset_properties}
\end{table*}

\subsubsection{\textbf{Hamiltonian prediction models}}

Models which train/predict Hamiltonian matrices were initially constrained to small-size systems, with most existing implementations treating small-molecule datasets such as MD17 \cite{Christensen2020} and QH9 \cite{qh9dataset}, or periodic materials datasets which are described by small, repeated unit cells \cite{deeph}. In these cases, rotational equivariance was satisfied by using tensor products that mix features \cite{deeph3, quantumham}. However, tensor product operations scale poorly with $L_{max}$, the maximum angular momentum degree of the spherical harmonic basis kept in the model, \(\mathcal{O}(L_{max}^6)\) when dense, and \(\mathcal{O}(L_{max}^5)\) when leveraging sparsity in the tensor coefficients \cite{priceoffreedom}. Despite significant acceleration with dedicated libraries \cite{better_cuet}, the computational complexity of tensor products explode when treating orbital interactions of \(L_{max}\ge 4\), as typically encountered in electronic structure matrices. 

Hamiltonian prediction architectures capable of treating systems beyond small molecules have thus taken a different route: They use eSCN convolutions to mix features. By relaxing strict parity equivariance and implementing local bond rotations to reduce the dimensionality of the spherical interactions from SO(3) to SO(2), eSCN-based models enable rotationally equivariant operations built from linear layers, and thus scale with \(\mathcal{O}(L_{max}^3)\) \cite{so2}. Currently, all Hamiltonian prediction models capable of treating structures with 100+ atoms rely on this approach \cite{wanet, amorphous, helm, deeph2}. A few of these models have recently been trained on `universal' datasets, making them capable of predicting orbital interactions across a wide range of atomic elements \cite{wanet, helm}. Elsewhere, the scalability of eSCN convolutions has also been leveraged for force/energy prediction, and serves as the backbone operation behind models such as the `Universal Model for Atoms' (UMA) \cite{uma}.

We note that there exists a class of electronic structure prediction models targeting `effective' or `empirical tight-binding' Hamiltonians rather than \textit{ab initio} ones at the DFT-level \cite{deeptb, hamster}. These models typically scale to much larger system sizes, but at the expense of the structural and compositional universality made possible by a DFT-level basis.

\subsubsection{\textbf{Domain decomposition/distribution}}

The vast majority of distributed GNN applications in materials science are concerned with MLIPs. Many of them, e.g., DeepMD \cite{deepmd} and Allegro \cite{Kozinsky2023}, leverage the domain decomposition functionality of the LAMMPS molecular dynamics code \cite{LAMMPS}. LAMMPS relies on spatial decomposition of the domain into boxes, and has an additional functionality to modify the box-boundaries and balance the number of atoms within each box. During message passing iterations, ghost atoms are communicated through halo nodes/edges at the boundaries to aggregate messages onto each node. The communicated volume is determined by the connectivity of the graph, densely connected graphs incurring high communication overhead as ghost atoms must send messages across multiple layers. 

Directly partitioning the graph instead of the spatial domain appears as a promising approach to improve load balancing and more evenly distribute communication. This is the approach pursued by SevenNet \cite{Park2024, Batzner2022} and DistMLIP \cite{distmlip}, which serve as wrappers for existing MLIP architectures. SevenNet is integrated into LAMMPS, and its strong scaling abilities are primarily limited by reduced GPU utilization when distributed over many GPUs. DistMLIP \cite{distmlip} supports multi-GPU (but single-node) inference. The distribution in these applications generally occurs only over nodes (atoms) and not over edges, making them less suitable to Hamiltonian prediction models, where the number of graph edges far exceeds that of nodes.


\subsection{Hamiltonian Matrix Datasets}

All DFT-level datasets available today have necessitated at least one electronic-structure calculation per atomic system considered, but almost none of them stored the corresponding Hamiltonian matrices. In \textbf{Table II} we summarize the parameters of several datasets that indeed include Hamiltonian matrices. Note that the nature of the atomic elements as well as the DFT features (functional, localized basis set) do not change the computational problem at hand, but heavier elements typically require a larger basis, thus determining $L_{max}$. To benchmark the computational performance of MALOQ when training on large-scale electronic structure matrix data, we use data from the three representative datasets in bold in \textbf{Table II}: (1) $\nabla^2$DFT and its drug-like molecules with up to 50 atoms each, (2) custom `unsolvated electrolytes' and `metal organics' datasets derived from the Open Molecules 2025 4 million data with basis-dependent $L_{max}$ \cite{omol}, and (3) the 3,000-atom/structure amorphous HfO$_2$ dataset released by Ref.~\cite{amorphous}.

\section{Implementation and Optimization}
 \label{sec:implementation}

\begin{figure}[t]
  \centering
  \includegraphics[width=\linewidth]{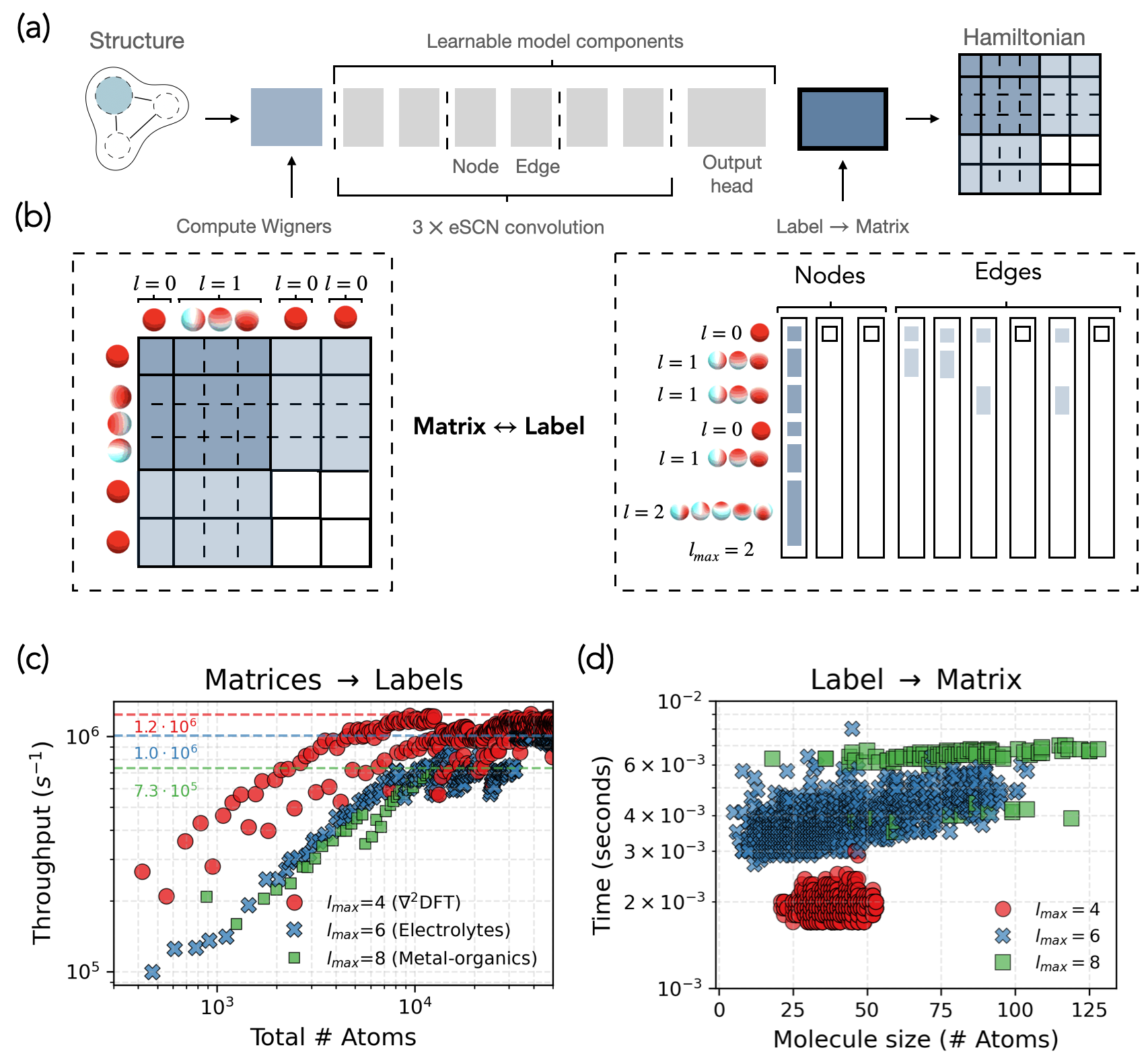}
  \caption{\textbf{Custom kernels for data processing and matrix reconstruction} (a) Schematic of the different steps along the inference process, highlighting (in blue boxes) the two non-learnable model components and in black frame the location where the `Matrix$\rightarrow$Label' conversion occurs. (b) Illustration showing the mapping between the matrix and label representations of data for a toy H$_2$O molecule with a minimal basis (Oxygen: $s$, $p$, Hydrogen: $s$), so that $L_{max}$=2. The labels for each node/edge are populated with diagonal/off-diagonal submatrices. Measured processing throughput/runtime to either (c) convert multiple matrices to labels, or (d) reconstruct a single matrix from a given set of predicted data. All measurements were taken over a subset of molecules present in the $\nabla^2$DFT ($L_{max}$=4) and OMol Electrolytes/Metal-Organics ($L_{max}$=6/8) datasets, with data processed in fp32.}
  \label{fig:matrix_to_label}
\end{figure}

MALOQ is an ML model for scalable prediction of electronic-structure operators, e.g., Hamiltonian and density matrices (rank-$N$), as well as general operator-valued quantities in materials modeling, such as Hessians (rank-2). We illustrate the stages of MALOQ's internal workflow during inference in
\textbf{Fig.~\ref{fig:matrix_to_label}(a)}. Overall, it consists of two components: a learnable ML model that learns mappings from atomic structure to spherical tensors, and `non-learnable' data-processing pipelines. 

The first `non-learnable' step, labeled `Compute Wigners', creates generalized rotation (Wigner-D) matrices which rotate all edges of the atomic graph to a common ($+y$) axis. These precomputed rotation matrices are then used within the second, learnable component of the model: the eSCN convolutions in each message passing layer. By rotating the embeddings of each node to a common axis before they interact, eSCN convolutions reduce the dimensionality of the `tensor product' between them from SO(3) to SO(2). The learnable interaction can then be applied through a set of constrained linear layers while preserving rotational equivariance, following a reference implementation first introduced in Ref.~\cite{so2}. Finally, the directed `messages' created by interacting nodes along each graph edge are either aggregated onto their target nodes (in the Node blocks) or used directly to update the embeddings of edges (in the Edge blocks).

At the end of the message passing operations, each node or edge of the atomic graph is described by an embedding which has been processed through three layers of learnable eSCN convolutions. At this point, the final `non-learnable' component of MALOQ ('Label$\rightarrow$Matrix in \textbf{Fig.~\ref{fig:matrix_to_label}(a)}) uses these predicted `Labels' to reconstruct the Hamiltonian in matrix representation. Note that during training, the opposite transformation is applied to the Hamiltonian matrices within the training dataset: 'Matrix$\rightarrow$Label performs basis transformations between the original, matrix representation of $\mathbf{H}$, and a set of labels which can be mapped to/learned by the individual nodes and edges of a GNN corresponding to the atomic structure of interest.

The non-learnable `data pipelines' within MALOQ have been accelerated through custom kernels that enable processing (training) over large dataset batches and inference for large system sizes. These large data volumes naturally exceed GPU memory constraints, necessitating a graph-level parallelization to process the corresponding embeddings through layers of eSCN convolutions. We thus additionally introduce a distributed-graph implementation which efficiently communicates embeddings across multiple processes, and illustrate partitioning strategies that reduce load imbalance across batches of molecules with diverse size, and large, single-materials graphs. Together, these computational contributions enable both rapid, scalable operator prediction, bringing the accuracy of MALOQ's backbone architecture \cite{helm} to 10k+ atom systems. In this section, we detail each of these components, starting with the methods behind them, associated computational challenges, and how they are optimized within the MALOQ code. We close each subsection with a brief evaluation of the computational improvements achieved.

\subsection{Training Environment}

We first introduce our experimental setup. Our models are trained on the Alps supercomputer at CSCS, where each node comprises 4 NVIDIA GH200 Grace Hopper superchips~\cite{hoefler_alps_bench}. Every superchip combines a Hopper GPU with 132 Streaming Multiprocessors and 96 GB HBM3 memory with a Grace CPU with 72 cores and 128 GB LPDDR5.
Each Hopper GPU is connected to every other GPU within the same node via NVLink and 150 GB/s bidirectional point-to-point bandwidth, resulting in a total of 900 GB/s.
The Alps nodes are connected via an HPE Slingshot 11 network with a Dragonfly topology.
MALOQ is written in PyTorch \cite{pytorch2024}, using torch tensors to represent the embeddings of nodes and edges. To communicate embeddings between GPUs, we use PyTorch's distributed functionality and its NVIDIA Collective Communications Library (NCCL) backend.

\subsection{Data pipelines: Matrix $\leftrightarrow$ Label conversions}

\subsubsection{\textbf{Methods}}

The entries of $\mathbf{H}$ correspond to interactions between electronic orbitals located on the same or on different atoms. The full matrix can be decomposed into sub-matrices ${\mathbf{H}^{\alpha\beta}_{ij}} = \bra{l_{\alpha}, m_{\alpha}}\mathbf{H}\ket{l_{\beta},m_{\beta}}$. They contain all interactions ${\mathbf{H}^{\alpha\beta}_{ij}}(m_{\alpha},m_{\beta})$ between spherical harmonic basis functions $l_\alpha$ (of degree $\alpha$) on atom $i$, and $l_\beta$ (of degree $\beta$) on $j$, as illustrated in the left part of \textbf{Fig.~\ref{fig:matrix_to_label}(b)}. Mathematically, each ${\mathbf{H}^{\alpha\beta}_{ij}}$ corresponds to the tensor product $l_\alpha \otimes l_\beta$, which has length $(2l_\alpha + 1)\times (2l_\beta + 1)$. Through a transformation $T$, this tensor product can be written as the direct sum ($\oplus$) of angular momentum eigenstates $\ket{L, M}$, where $L$ ranges from degree $|\alpha-\beta|$ to $|\alpha+\beta|$, i.e., $T(l_\alpha \otimes l_\beta)  = \bigoplus_{i=|\alpha - \beta|}^{(\alpha + \beta)}L_i = L_{|\alpha-\beta|} \oplus...\oplus L_{(\alpha+\beta)}$. The transformation $T$ involves a matrix of Clebsch-Gordon (CG) coefficients. The entries for a specific $\ket{L, M}$ component are given by:

\vspace{-0.1in}
\begin{equation}
\ket{L, M} = \sum_{m_\alpha=-l_\alpha}^{+l_\alpha}\sum_{m_\beta=-l_\beta}^{+l_\beta}C_{(l_\alpha, m_\alpha)(l_\beta, m_\beta)}^{(L, M)}\ket{l_\alpha, m_\alpha}\ket{l_\beta, m_\beta}.
\label{eqn:CG}
\end{equation}

\noindent Here, $C_{(l_\alpha, m_\alpha)(l_\beta, m_\beta)}^{(L, M)}$ is the CG coefficient describing the contribution of $\ket{l_\alpha, m_\alpha}\ket{l_\beta, m_\beta}$ to the state $\ket{L, M}$, and $\ket{l_\alpha, m_\alpha}\ket{l_\beta, m_\beta}$ corresponds to ${\mathbf{H}^{\alpha\beta}_{ij}}(m_{\alpha},m_{\beta})$. The transformation $T$ thus decomposes the Hamiltonian sub-matrix $\smash{\mathbf{H}^{\alpha\beta}_{ij}}$ into a `vector' form, representing a direct sum of tensors with different angular degree $l$. All $\ket{L, M}$ corresponding to a single ${\mathbf{H}_{ij}}$ block are then concatenated into a vector $\mathbf{h}_{ij}$. The same operation is repeated for all blocks of $\mathbf{H}$. Since not all atomic elements require the same number of orbitals (basis set) to be accurately described, the resulting $\mathbf{h}_{ij}$ vectors have different lengths. They are all padded to a common dimension, that of the largest element in the dataset. The collection of all padded $\mathbf{h}_{ij}$ forms the `labels' used to train the node and edge embeddings of the GNN (right part of \textbf{Fig.~\ref{fig:matrix_to_label}(b)}), and the loss is computed element-wise over the model's prediction of them. The maximum degree of the $\mathbf{h}_{ij}$ becomes the $L_{max}$ required for embeddings.


\subsubsection{\textbf{Optimization}}

To process a dataset of Hamiltonian matrices, we need to (1) extract the individual orbital interaction blocks into equally-shaped buffers and (2) perform a basis transformation to convert them into labels using \textbf{Eq.~\eqref{eqn:CG}}. While the matrix operations within the basis transformation step can be done efficiently using PyTorch tensors, the coefficient extraction in \textbf{Fig.~\ref{fig:matrix_to_label}(b)} rapidly becomes a bottleneck for larger training datasets or inference system sizes.

We introduce a dedicated CUDA kernel to handle the transformations between batches of (or single) matrices and labels. Within this kernel, each thread processes a single matrix element, i.e., orbital-orbital interaction block, and each thread block processes one node (${\mathbf{H}^{\alpha\beta}_{ii}}$) or edge (${\mathbf{H}^{\alpha\beta}_{ij}}$). The kernel uses an `orbital template' $\mathcal{T}(Z_i, Z_j)$ containing the row and column slices of $\mathbf{H}_{ij}$ corresponding to the set of orbitals interaction between atom $i$ of atomic element $Z_i$ and atom $j$ of atomic element $Z_j$. This information can be precomputed for a given basis. 

As the size of the labels differs for each dataset, we benchmark our kernel on a selection of three molecular datasets in \textbf{Table \ref{tab:dataset_properties}}, each with different $L_{max}$ and orbital basis.
To evaluate the performance under realistic conditions, we consider two different scenarios: (i) data processing (forward transformation, pre-training) and (ii) matrix reconstruction (backward transformation, post-inference).
First, \textbf{Fig.~\ref{fig:matrix_to_label}(c)} reports the `data processing' case (`Matrices $\rightarrow$ Labels'), where multiple Hamiltonian matrices $\mathbf{H}$ are converted into orbital interaction labels $\mathbf{h}$ for supervised learning. We evaluate processing throughput by considering increasing fractions of each of the three datasets, resulting in increased total \# atoms. Processing throughput between 7.3$\times$10$^{5}$ and 1.2$\times$10$^{6}$ atoms/s are measured, allowing us to process the largest molecular datasets of 110M structures \cite{omol} ($\sim$50 atoms/structure) for training in under an hour, on a single process. Note that for $L_{max}=8$, the data exceeds the memory of a single GH200 GPU before peak processing throughput can be reached. We therefore report the maximum value attained. Secondly, in \textbf{Fig.~\ref{fig:matrix_to_label}(d)}, the `matrix reconstruction` case is presented (`Labels $\rightarrow$ Matrix', $\mathbf{h}\rightarrow\mathbf{H}$). At the end of the inference step, the model generates orbital interaction labels $\mathbf{h}$ for a given atomic structure, which are then re-assembled into the $\mathbf{H}$ for downstream applications. Here, the difference in runtime stems primarily from the number of elements per label ($L_{max}$), but remains under 10 ms across a range of system sizes. Note that the variability in measured time/throughput at the same \# atoms can be attributed to different molecular connectivity (\# edges per atomic graph).

\begin{figure}[t]
    \centering
    \includegraphics[width=\linewidth]{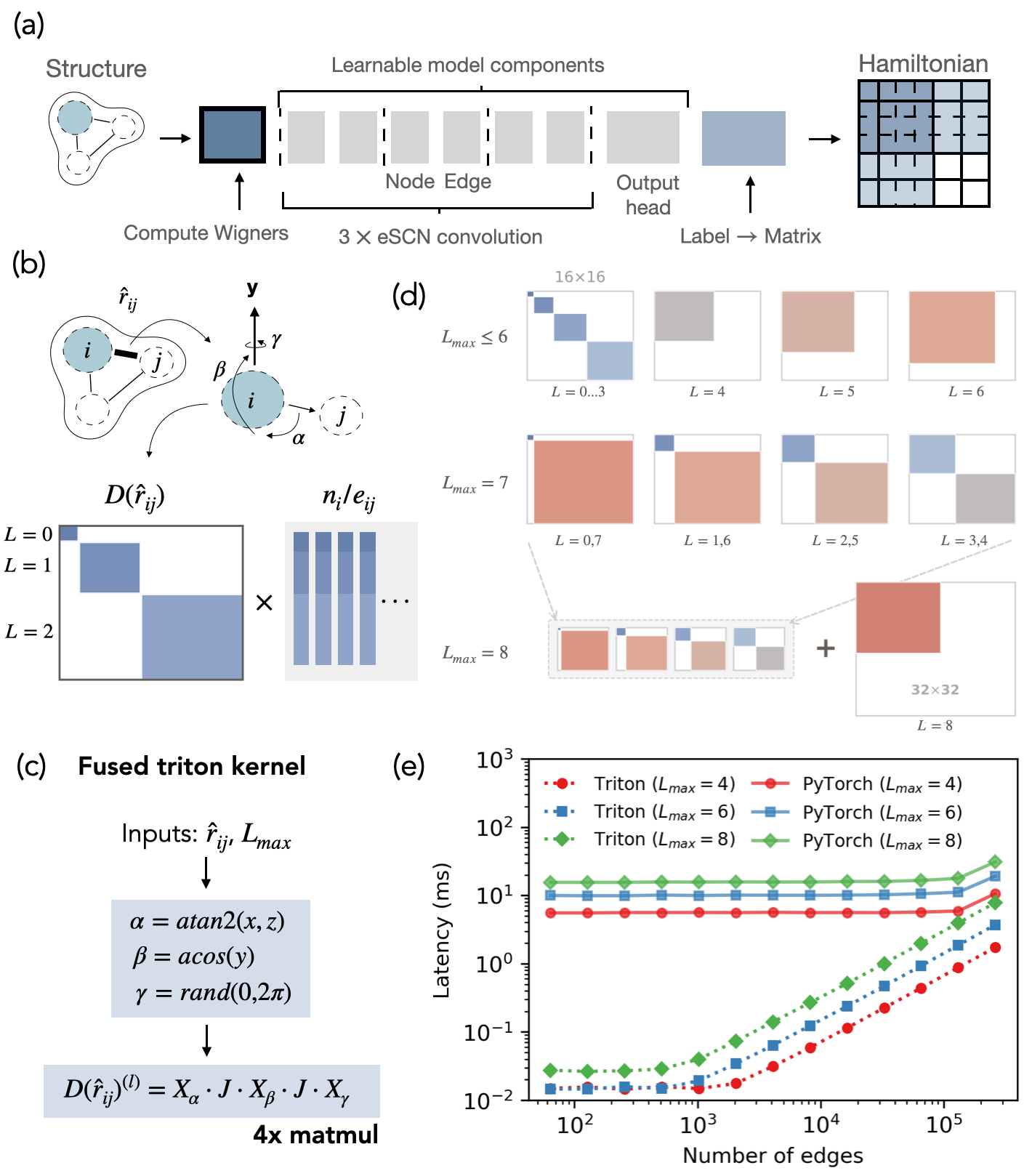}
    \caption{\textbf{Fused Triton kernel for Wigner-$D$ matrix construction.} (a) Schematic of the inference process highlighting the location where rotations occur in the MALOQ application (blue box with thick black frame). (b) These Wigner-D matrices rotate the node/edge embedding $n_i/e_{ij}$ of every edge $\hat{r}_{ij}$ from the atomic graph to align it with the $+y$-axis. (c) Doing this requires first computing the Euler angles ($\alpha, \beta, \gamma$) of the rotation from edge vectors $\hat{r}_{ij}$, and then constructing $D^{(\ell)}$ via four tl.dot operations, which we encapsulate into a fused Triton kernel. (d) Block-packing strategy for matrix product (Triton tl.dot): $\ell$-blocks are packed into $16{\times}16$ tiles (or $32{\times}32$ for $\ell{=}8$) to maximize tensor core utilization. (e) Latency comparison between our Triton kernel and the Torch implementation at selected number of input edges.}
    \label{fig:wigner}
\end{figure}

\subsection{Data pipelines: Accelerated Wigner-$D$ matrix generation}

\subsubsection{\textbf{Methods}}

We now turn to the Wigner-D matrix block in \textbf{Fig.~\ref{fig:wigner}(a)}. For a given input atomic graph, MALOQ first pre-computes the set of generalized rotation matrices $R$ that rotate the graph edges ($e_{ij}$) to a common ($+y$) axis (\textbf{Fig.~\ref{fig:wigner}(b), top}). As in many equivariant GNNs for materials science, the embeddings of every node ($n_i$) and edge ($e_{ij}$) often take the structure of spherical harmonic coefficients of rank $L=0...L_{max}$. Each $L$-coefficient is expanded into $E$ channels - the dimensions of every node ($n_i$) and edge ($e_{ij}$) embedding processed is thus $(L_{max}+1)^2\times E$. Performing a rotation \( R \) of the original atomic system $\hat{r}$ into $\hat{r}'=R\cdot \hat{r}$ (where \( \hat{r} \)/\( \hat{r}' \) are normalized direction vectors) requires the embeddings  $n_i$/$e_{ij}$ to be multiplied by a direct sum of `Wigner-D' matrix $D^{L}(R)$ of degree $L$., i.e., $n_i'$/$e_{ij}' = \bigoplus_{l=0}^{L_{max}}D^{l}(R)\times n_i$/$e_{ij}^l$. The lower part of \textbf{Fig.~\ref{fig:wigner}(b)} illustrates the block-diagonal structure of these Wigner-D matrices which rotate the corresponding embeddings $n_i$/$e_{ij}$. Practically, creating the Wigner-D matrices involves (1) computing a set of Euler angles $\alpha, \beta, \gamma$ corresponding to each edge in a given atomic graph, and then (2) using them to determine the (2$L$+1$)\times($2$L$+1) entries of $D^{L}(R)$.

\subsubsection{\textbf{Optimization}}

A reference PyTorch implementation is provided by the e3nn library~\cite{e3nn}, which computes each $D(R)^L$ submatrix separately, but for all graph edges in a batch.
Each sub-computation involves a chain of four small matrix multiplications, executed via separate kernel launches, materializing the intermediate results in GPU global memory.
Thus, a significant speedup can be gained by developing a fused kernel to perform the full \( \hat{r}_{ij} \) $\rightarrow$ $D(R)^L$ operation chain for every graph edge $\hat{r}_{ij}$ and orbital degree $l$ (\textbf{Fig.~\ref{fig:wigner}(c)}), vastly increasing arithmetic intensity.
This kernel, which we implement in Triton, (1) fuses the calculation of Euler angles and of all matrix products that are required to obtain the Wigner-D sub-matrices for every degree $L...L_{max}$, and (2) implements a block-packing strategy to group different $L$s (\textbf{Fig.~\ref{fig:wigner}(d)}). For the latter, we take advantage of the regular angular dimensions of each $L$, which are fixed by the underlying physics: The size of each sub-matrix $D^{l}$ is (2$l$+1)$\times($2$l$+1), to account for the coefficients $m = -l...+l$ for a spherical harmonic of degree $l$. As the matrix product in Triton (tl.dot) uses fixed shapes (e.g., 16$\times$16, 32$\times$32...), packing the sub-matrices for multiple $L$-components together minimizes padding overhead and fully utilizes the fixed-size tiles.

We benchmark the reference PyTorch and our Triton implementations with three different $L_{max}$ values, corresponding to selected datasets from \textbf{Table~\ref{tab:dataset_properties}}.
The results are presented in \textbf{Fig.~\ref{fig:wigner}(e)}.
At smaller data sizes (number of edges, and $L_{max}$), the execution time in the PyTorch reference implementation is dominated by kernel launch overhead from multiple matrix products. In this range, our Triton code shows speedups of $\sim$500$\times$ (100-1K edges) to $\sim$50$\times$ (10K edges), as the launching overhead was removed through kernel fusion. At higher data volumes (100K+ edges), the reference implementation becomes limited by global memory traffic. Still, our Triton implementation achieves a $\sim$5-6$\times$ speedup from the combination of kernel fusion and $L$-wise block packing.

\begin{figure}[t]
  \centering
  \includegraphics[width=0.95\linewidth]{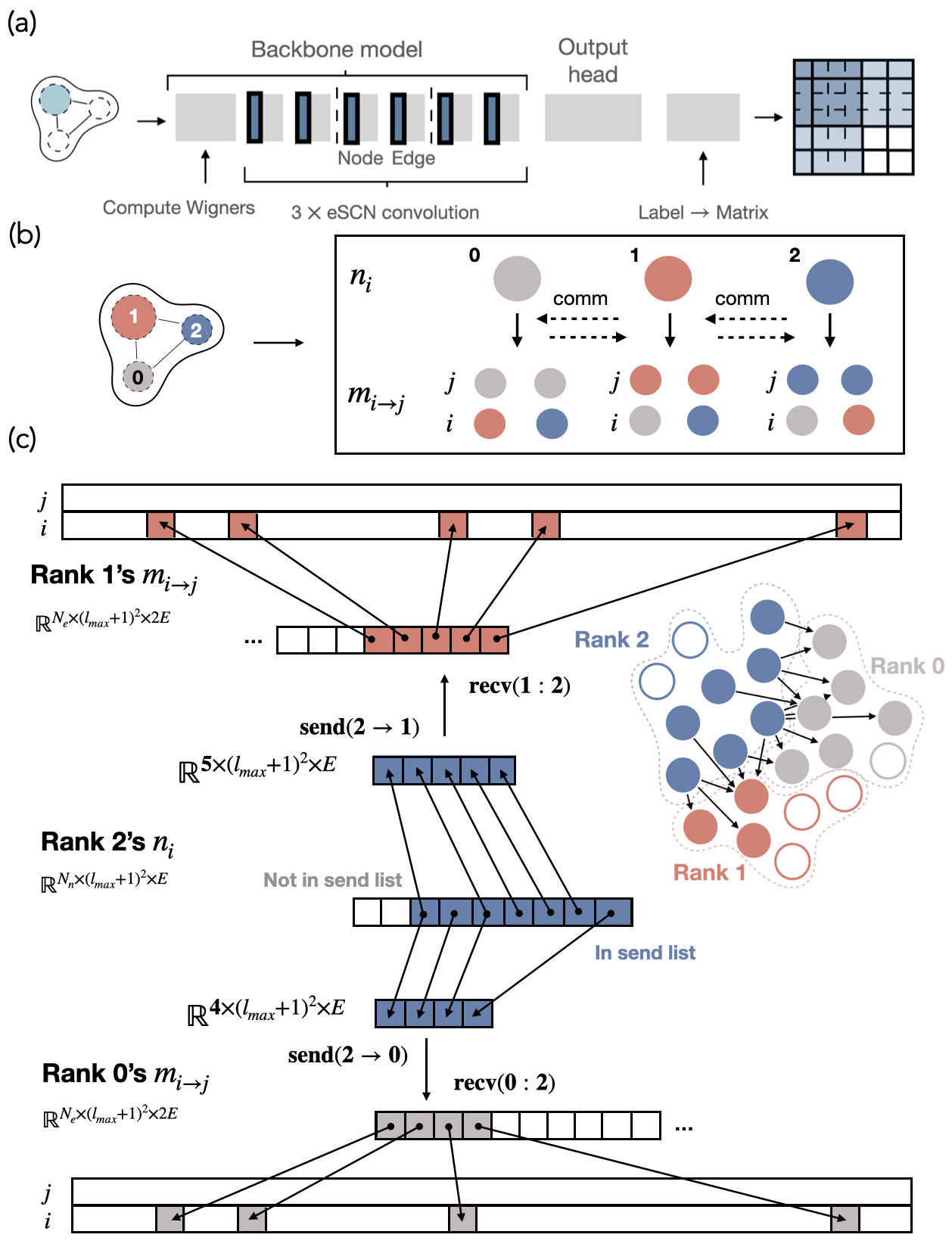}
  \caption{\textbf{Custom communication scheme for graph-level distribution.} (a) Schematic of the inference process showing where the communication occurs within MALOQ, i.e., at the start of each node/edge block within a eSCN convolution layer (black frames). (b) Illustration of the communication pattern required to assemble the messages $m_{i\rightarrow j}$ within a toy H$_2$O molecule. (c) Illustration of the send/recv operation implementation to communicate node embeddings from Rank 2 to Ranks 0 and 1 for the example partitioned graph pictured in the inset. The arrows indicate the edges starting from nodes belonging to Rank 2's partition and ending in one of the other partitions: 4 (5) Rank 2's nodes are connected to Rank 0 (1)'s nodes through inter-partition edges. The corresponding data is first copied to a buffer before being sent to the targeted Rank. There, it is unpacked into a local message buffer.}
\label{fig:partitioning}
\end{figure}

\subsection{Graph level distribution}


At the start of each MP layer (indicated with black frames in \textbf{Fig.~\ref{fig:partitioning}(a)}), a set of messages $m_{i\rightarrow j}$, consisting of the concatenated node embeddings $n_i \oplus n_j$, are assembled. \textbf{Figure~\ref{fig:partitioning}(b)} illustrates this process for a fully-connected toy molecule of 3 atoms, picturing the construction of each of the 6 messages required.
This set of messages undergoes eSCN convolutions, which mix the features of $n_i$ and $n_j$, and are then either aggregated onto node $j$ (node update) or used to update the edge embedding $e_{ij}$ (edge update).
In a distributed-memory setting, constructing these messages requires communication of node embeddings across graph partitions.
Our implementation uses PyTorch's distributed NCCL backend and aggregates the communication required between any two processes as follows:
Each process
(1) packs its own $n_i$ into a separate send buffer for every one of its neighbors ($j$),
(2) sends/receives those buffers using point-to-point communication,
and (3) unpacks the received $n_i$ and inserts them into its message tensors.
This communication scheme is visualized in \textbf{Fig.~\ref{fig:partitioning}(c)}, where a process (Rank 2) packs and sends its local $n_i$ required by its neighbors (Ranks 0 and 1), which receive and insert them into their $m_{i\rightarrow j}$.
During the backward pass, the reverse communication pattern is executed, implemented as a custom function in PyTorch.

In addition, we make several design choices to further refine the communication step.
First, we adopt a target-nodes-own-edges distribution policy:
Each rank owns the edges $e_{ij}$ where the connected node $j$ is local, eliminating additional inter-process communication during the edge update phase. In addition, we partially hide communication latency by overlapping the indexing of local nodes into message tensors with the asynchronous transfer of remote node embeddings. To that end, we partition the list of local node embeddings such that those requiring transmission to remote ranks are contiguous in memory, as shown for Rank 2 in \textbf{Fig.~\ref{fig:partitioning}(c)}.

The remaining objective is to partition the graph in order to evenly distribute the workload and reduce communication time. In the following, we discuss and compare the different graph partitioning approaches implemented in MALOQ for applications across molecules and materials.

\begin{figure}[t]
  \centering
  \includegraphics[width=\linewidth]{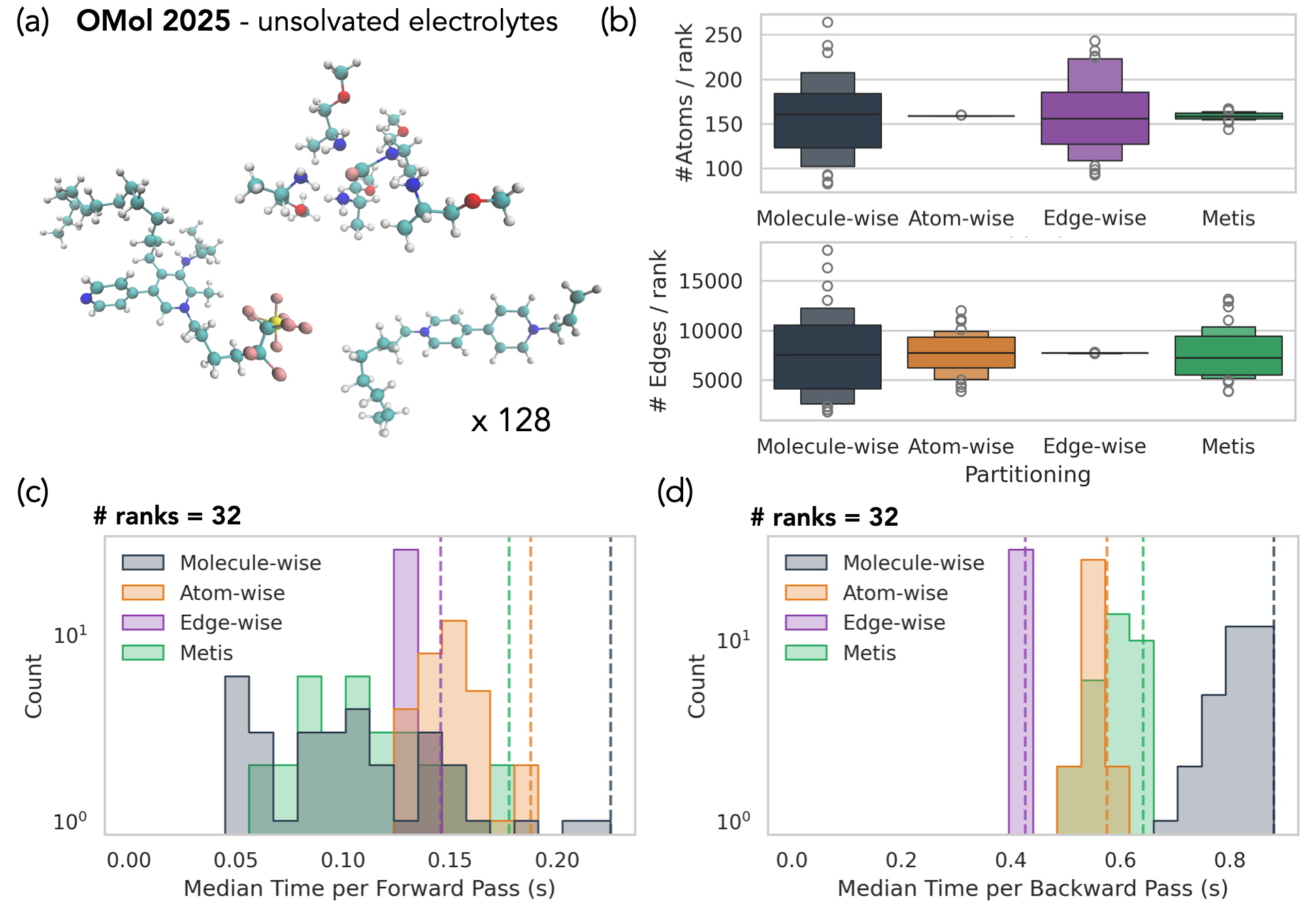}
  \caption{\textbf{Training runtime comparison under different distribution schemes.} (a) Dataset used for training. It consists of 128 randomly-selected molecules from the OMol electrolytes dataset and their corresponding $\mathbf{H}$. (b) Workload distribution across ranks quantified in \# atoms and \# edges, for three different distribution schemes (atom-wise, edge-wise, Metis), compared to the undistributed case in which 128 molecules are divided across 32 ranks (molecule-wise). (c) Forward and (d) backward pass time distribution across different ranks. Each data is the median of 200 training epochs, of which the first 20 are discarded. The vertical dashed lines indicate the maximum time for each distribution scheme.}
  \label{fig:omol_dist_32rank}
\end{figure}

\subsubsection{\textbf{Partitioning molecular datasets}}
Training on molecular datasets in a distributed-memory environment typically involves creating a molecule-wise partition, where each rank processes a subset of molecules.
Load imbalance might arise if these molecules are made of a wide range of atoms, which may contain an even wider distribution of connectivity (\# edges per structure). This load imbalance can be reduced by `binning' molecules into partitions to balance the total number of atoms/edges across the structures handled by each process.
However, a finer load balance can be achieved by partitioning the molecules at the atom/edge level. In practice, this can significantly reduce the time to process each training batch.

\begin{figure}[t]
  \centering
  \includegraphics[width=\linewidth]{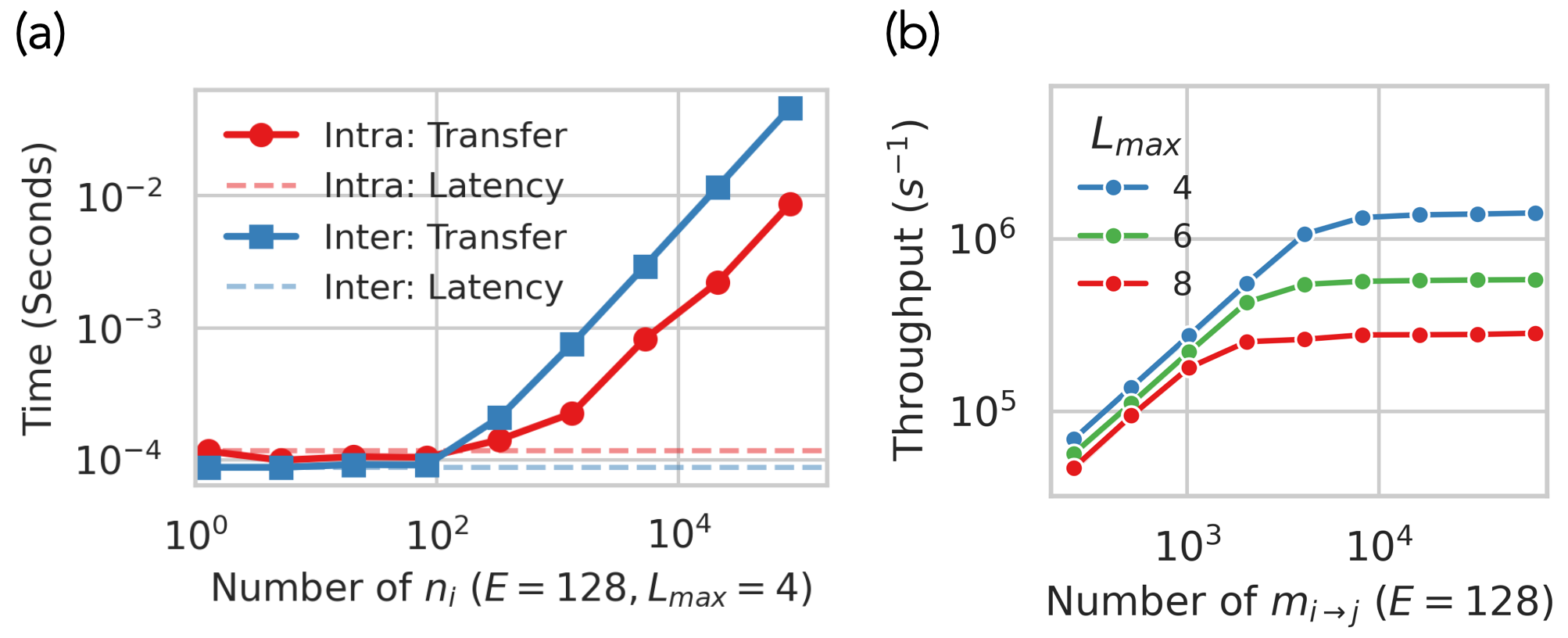}
  \caption{\textbf{Embedding communication and processing throughput}. (a) Measured intra-/inter-node communication throughput of node embeddings $n_i$ with embedding size $E$=128. (b) Processing throughput for eSCN convolutions as a function of the number of messages $m_{i\rightarrow j}$ for different $L_{max}$. Each data point corresponds to the median of 20 measurements.}
  \label{fig:batch_throughput}
\end{figure}

\begin{figure*}[t]
  \centering
  \includegraphics[width=\linewidth]{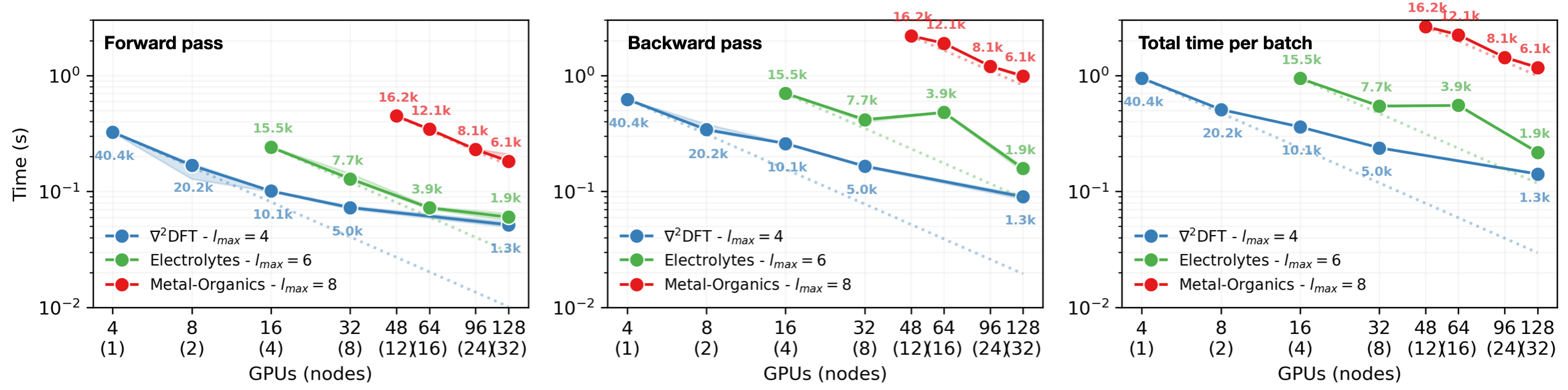}
  \caption{\textbf{Strong scaling of MALOQ over three irregular molecular datasets from \textbf{Table~\ref{tab:dataset_properties}} with different $L_{max}$.} (a) Forward pass. (b) Backward pass. (c) Total time per batch. A total of 128 molecules from each dataset is placed in each batch, which is distributed according to an edge-wise partition. The data plotted is the median over 200 epochs (the first 20 are discarded), and the shaded areas show variance over runtime across ranks. Each data point is annotated by the average number of edges owned by each process under that distribution, to compare with the saturation of embedding processing throughput measured in \textbf{Fig.~\ref{fig:batch_throughput}}. Note that we start from the minimum number of GPUs allowing for the partitioned batch to fit in memory.}
  \label{fig:molecule_strongscaling}
\end{figure*}

As an example, we consider in \textbf{Fig.~\ref{fig:omol_dist_32rank}} a batch of 128 molecules from the OMol-electrolytes dataset (\textbf{Fig.~\ref{fig:omol_dist_32rank}(a)}) and compare the workload distribution (\textbf{Fig.~\ref{fig:omol_dist_32rank}(b)}), as well as the runtime per forward (\textbf{Fig.~\ref{fig:omol_dist_32rank}(c)}) and backward pass (\textbf{Fig.~\ref{fig:omol_dist_32rank}(d)}) under four different partitioning schemes. The so-called `Metis' approach uses the METIS package~\cite{metis} to search for a mincut solution to the partitioning problem. Since cuts to the graph are penalized, Metis finds an `optimal binning' when there are more molecules than ranks, thus balancing the partitions without introducing inter-process communication. This is the case in~\textbf{Fig.~\ref{fig:omol_dist_32rank}(b)}. Meanwhile, in the `atom-wise'/`edge-wise' partitions, nodes/edges are evenly distributed among ranks, starting from a list of atoms across the data batch where those within each molecule are contiguous. In particular, an edge-wise distribution of the graph leads to a highly balanced embedding workload. Compared to the molecule-wise distribution, splitting the batch edge-wise reduces the time per forward pass by 30\% (from 0.23 to 0.15s), and the time per backwards pass by over 50\% (0.85 to 0.4s), despite the introduction of inter-process communication.

\subsubsection{\textbf{Partitioning large materials graphs}}

In case of large atomic structures, the corresponding materials graph must be distributed over multiple processes. The resulting communication involves only a few embeddings exchanged between any two ranks, and is eventually limited entirely by overhead. To highlight this effect, we show in \textbf{Fig.~\ref{fig:batch_throughput}(a)} the time to send/receive different counts of $n_i$ between two processes, each running on a separate GH200 superchip on Alps.
Up to about 100 $n_i$ (considering $E=128$ channels and $L_{max}=4$), communication is bound by latency. At the same time, the achievable runtime improvements due to graph distribution are limited by processing throughput, which decreases once each partition owns too few local edges. The number of messages required to saturate processing throughput is shown in \textbf{Fig.~\ref{fig:batch_throughput}(b)}. It is proportional to the number of local edges.


To minimize the number of partition neighbors and data exchanges in large materials graphs (and thus the latency), we introduce an additional partitioning method based on recursive bisection. Our `modified recursive bisection' scheme recursively distributes a graph among 2$^n$ ranks while maintaining edge-wise balanced partitions, accounting for 3-D periodicity used in DFT calculations of materials, and minimizing the number of new neighbors introduced by each local cut. Starting from standard recursive bisection, we make the following changes: \textbf{(1)} The dimension along which to cut is determined based on ‘which dimension would create the fewest additional neighboring partitions when cut’, considering both (a) the interaction cutoff radius $r_{cut}$ and (b) periodicity of the 3-D domain. \textbf{(2)} Each sub-domain, at any stage of the algorithm, is partitioned so that the total node degree (N$_D$), which we use as a proxy for the number of edges, remains balanced between the two newly formed partitions.

\section{Applications}

MALOQ is designed to \textbf{(1)} improve parallel efficiency (time per epoch) when training on massive (terabyte-scale) Hamiltonian matrix datasets and \textbf{(2)} rapidly infer electronic structure matrices for materials at scale. In this section, we present performance benchmarks for real training scenarios under both operating conditions, using the datasets from \textbf{Table II}. These benchmarks are specific to the hardware discussed in \textbf{Section III-A} - other architectures with different memory constraints are likely to show different scaling performance.

\subsection{Scalable training over large molecular datasets}

`Universal' molecular datasets with diverse atomic elements typically necessitate large batch sizes during training, to ensure sufficient sampling over all of the represented elements within each batch.
We thus standardize all our benchmarks using a single batch size of 128 randomly selected molecules, which is distributed across multiple processes.
In all cases, we use an edge-wise partitioning, as previously explored in \textbf{Fig.~\ref{fig:omol_dist_32rank}}. Note that tackling the largest molecules (in terms of \# atoms, $L_{max}$) across these datasets already requires partitioning individual molecular graphs to fit in GPU memory.


The strong scaling results for the training runtime of three molecular datasets are plotted in \textbf{Fig.~\ref{fig:molecule_strongscaling}}: $\nabla^2$DFT, electrolytes, and metal-organics (Table~\ref{tab:dataset_properties}).
Each dataset necessitates a different $L_{max}$, up to 8 for metal-organics which contain lanthanide elements with $g$-orbitals.
We train these datasets on up to 32 Alps nodes (128 GH200 superchips), where the minimum number of nodes depends on the dataset: $\nabla^2$DFT's batch fits on a single node, electrolytes require at least 4 nodes, and metal-organics cannot run on less than 12 nodes.
Particularly in the forward pass, the number of edges per GPU at which the scaling efficiency starts dropping roughly corresponds to the point where processing throughput is lost in \textbf{Fig.~\ref{fig:batch_throughput}(b)}: $\sim$10k for $\nabla^2$DFT with $L_{max}=4$, and $\sim$5k for the electrolytes with $L_{max}=6$. 
The times for the metal-organics molecules scale well up to 32 nodes, since the average number of edges per partition (6.1k) remains well above the low saturation point for $L_{max}=8$ (about 2k). 


\subsection{Training and inference over large materials}

\begin{figure}[t]
  \centering
  \includegraphics[width=0.8\linewidth]{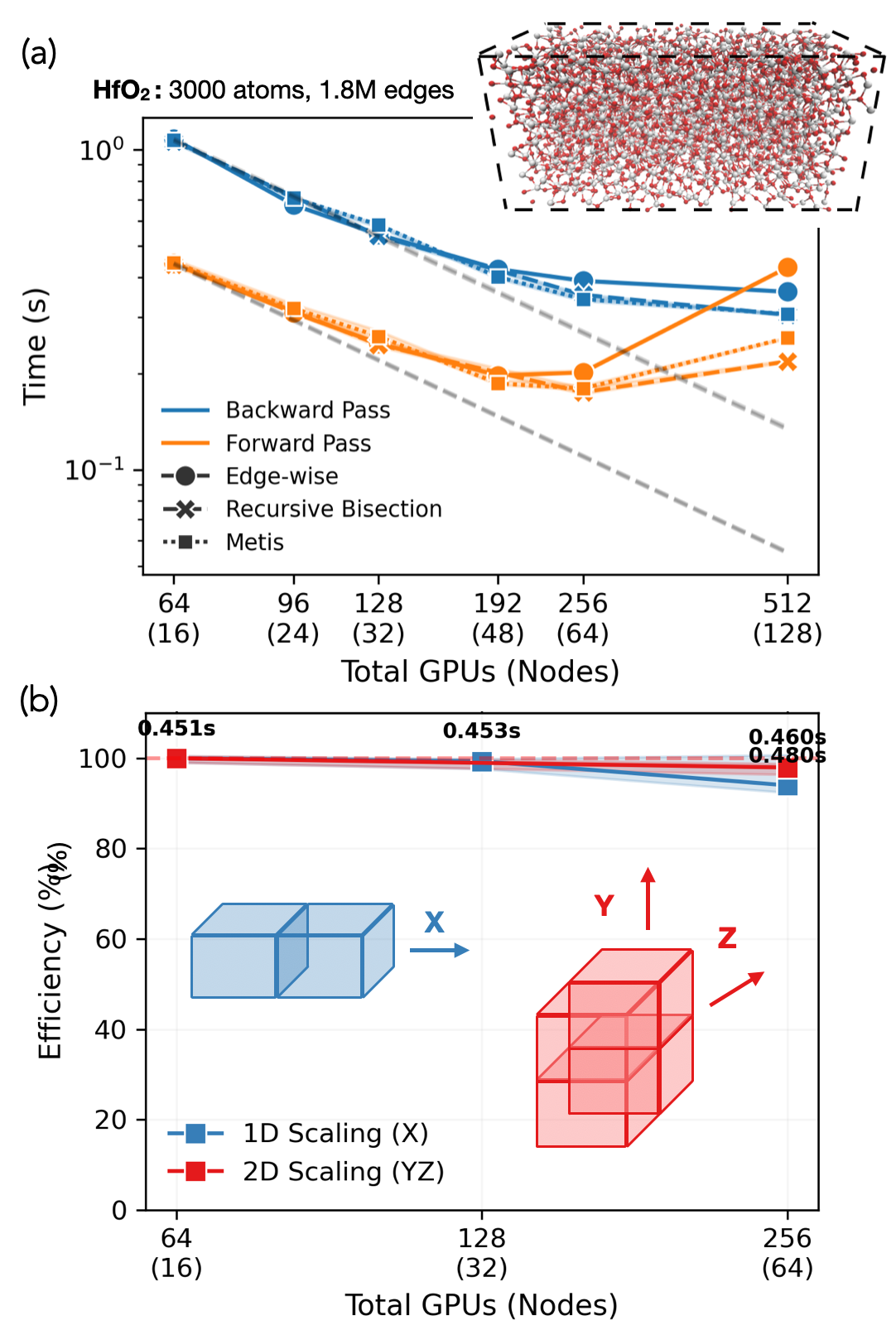}
  \caption{\textbf{Scalable training and inference on large materials.} (a) Strong scaling of MALOQ's training on large material datasets. The times for the forward (orange) and backward (blue) passes are reported for three different partitioning of an amorphous 3000-atom HfO$_2$ structure (shown in inset). (b) Weak scaling over inference for the same structure as in (a). The problem size is increased according to two tiling scheme: 1-D (tiling along $x$) and 2-D (tiling along $y$ and $z$). We measure 92.4\% scaling efficiency between 64 and 256 GPUs in the former case, 96.4\% in the latter. For (a-b), the reported measurements represent the per-rank median over 180 epochs, after 20 warm-up epochs. Each data point also contains a range (fill-between) indicating the variability in runtime across ranks.}
  \label{fig:hfo2_weakscaling}
\end{figure}

Describing material phenomena like phase boundaries \cite{Zhou2023}, interfaces between different materials \cite{Dossena2025}, or amorphous structures \cite{amorphous}, translates to the construction of large material graphs containing 10k+ atoms. When processing node and edge embeddings for structures at these scales, taking advantage of the memory of multiple GPUs through graph-level distribution becomes necessary.

First, we test the effect of different partitioning schemes when training on large material graphs. We report in \textbf{Fig.~\ref{fig:hfo2_weakscaling}(a)} the strong scaling performance of MALOQ for the HfO$_2$ material dataset from \textbf{Table II}. Here, a single structure, corresponding to one batch, contains 3,000 atoms, and roughly 1.8 million edges under a cutoff $r_{cut}=6.0$ \AA. We measure strong scaling efficiency up to 512 GPUs (128 nodes) on the Alps supercomputer - at this distribution, each rank has, on average $\sim3\times10^{3}$ edges. This is the maximum `reasonable' distribution for this problem size, as further distribution would reduce the number of edges per rank, with loss of processing throughput as a consequence (see \textbf{Fig.~\ref{fig:batch_throughput}(b)}). We compare three partitioning strategies: \textbf{(1)} the same edge-wise distribution as for molecules, now performed on the material graph sorted by $x$-coordinates, \textbf{(2)} the Metis algorithm, and \textbf{(3)} our modified recursive bisection. Partitioning with Metis significantly improves strong scaling over a straightforward edge-wise distribution, as it preserves clusters of locality in such 3-D material graphs. The modified recursive bisection approach slightly outperforms Metis on 512 GPUs, where significant communication times amplify the impact of the partitioning method. In general, Metis and our recursive bisection approach both enable scaling up to 512 GPUs for the backward pass and up to 192 GPUs for the forward pass.

\textbf{Figure~\ref{fig:hfo2_weakscaling}(b)} presents weak scaling efficiency for inference using the same amorphous-HfO$_2$ system.
We start on 16 Alps nodes (64 GH200 superchips) with the baseline structure made of 3,000 atoms ($\sim$1.8 million edges), and scale up to 64 nodes and 12,000 atoms ($\sim$10 million edges). To reliably increase the problem size, we tile the initial 3,000-atom HfO$_2$ cell according to two strategies: \textbf{(1)} 1-D tiling, where the cell is repeated only along the $x$-direction, and \textbf{(2)} 2-D tiling, where the cell is repeated along the $y$- and $z$-dimensions. In both cases, we use our modified recursive bisection algorithm to partition the tiled graphs. MALOQ achieves nearly ideal weak-scaling efficiency for inference on these structures, 94\% with 1-D scaling and 98\% with 2-D scaling.


\section{Conclusion and Outlook}

We introduced MALOQ, a scalable, distributed ML operator prediction code capable of treating higher-rank tensor properties, such as Hessians (rank-2) \cite{hip}, and electronic structure (Hamiltonian/Density) matrices where the rank ($L_{max}$) is determined by the orbital basis of the underlying data \cite{helm, wanet, Nigam2022, deeph3}. It shares its backbone architecture (distributed equivariant GNN with eSCN convolutions) with common energy/force property prediction models, thus allowing for distributed pre-training and finetuning across different quantities. MALOQ can be combined with advanced quantum transport solvers to reveal key features of electronic devices requiring large atomic structures, such as phase-change memory \cite{Zhou2023}, silicon nano-ribbon field-effect transistors \cite{Yeap2024}, or nano-ionic resistive memories \cite{Onofrio2015, Kaniselvan2023}.

\section*{Acknowledgment}
This work was supported by the Swiss National Science Foundation (SNSF) under grant $\mathrm{n^\circ}$ 209358 (QuaTrEx), and by the Platform for Advanced Scientific Computing in Switzerland (BoostQT). We acknowledge support from CSCS (projects c33, lp16, lp82).
The authors would like to especially thank Tim Robinson (CSCS) for access to and support of the computational resources.

\bibliographystyle{IEEEtran}
\bibliography{biblio}

@article{Venturella2025,
  title = {Unified deep learning framework for many-body quantum chemistry via Green’s functions},
  ISSN = {2662-8457},
  url = {http://dx.doi.org/10.1038/s43588-025-00810-z},
  DOI = {10.1038/s43588-025-00810-z},
  journal = {Nature Computational Science},
  publisher = {Springer Science and Business Media LLC},
  author = {Venturella,  Christian and Li,  Jiachen and Hillenbrand,  Christopher and Leyva Peralta,  Ximena and Liu,  Jessica and Zhu,  Tianyu},
  year = {2025},
  month = jun 
}

@misc{Seongsu2026,
  doi = {10.48550/ARXIV.2602.16897},
  url = {https://doi.org/10.48550/ARXIV.2602.16897},
  author = {Kim,  Seongsu and Lee,  Chanhui and Kim,  Yoonho and Yun,  Seongjun and Kim,  Honghui and Kim,  Nayoung and Park,  Changyoung and Han,  Sehui and Lim,  Sungbin and Ahn,  Sungsoo},
  title = {Machine Learning Hamiltonians are Accurate Energy-Force Predictors},
  publisher = {arXiv},
  year = {2026},
  copyright = {Creative Commons Attribution 4.0 International}
}

@inproceedings{nablaDFT,
title={\${\textbackslash}nabla{\textasciicircum}2\${DFT}: A Universal Quantum Chemistry Dataset of Drug-Like Molecules and a Benchmark for Neural Network Potentials},
author={Kuzma Khrabrov and Anton Ber and Artem Tsypin and Konstantin Ushenin and Egor Rumiantsev and Alexander Telepov and Dmitry Protasov and Ilya Shenbin and Anton M. Alekseev and Mikhail Shirokikh and Sergey Nikolenko and Elena Tutubalina and Artur Kadurin},
booktitle={The Thirty-eight Conference on Neural Information Processing Systems Datasets and Benchmarks Track},
year={2024},
url={https://openreview.net/forum?id=ElUrNM9U8c}
}

@misc{orbitall,
  doi = {10.48550/ARXIV.2507.03853},
  url = {https://doi.org/10.48550/ARXIV.2507.03853},
  author = {Kang,  Beom Seok and Bhethanabotla,  Vignesh C. and Tavakoli,  Amin and Hanisch,  Maurice D. and Goddard,  William A. and Anandkumar,  Anima},
  title = {OrbitAll: A Unified Quantum Mechanical Representation Deep Learning Framework for All Molecular Systems},
  publisher = {arXiv},
  year = {2025},
  copyright = {Creative Commons Attribution 4.0 International}
}

@inproceedings{Vetsch2025,
  series = {SC ’25},
  title = {Ab-initio Quantum Transport with the GW Approximation,  42, 240 Atoms,  and Sustained Exascale Performance},
  url = {http://dx.doi.org/10.1145/3712285.3771784},
  DOI = {10.1145/3712285.3771784},
  booktitle = {Proceedings of the International Conference for High Performance Computing,  Networking,  Storage and Analysis},
  publisher = {ACM},
  author = {Vetsch,  Nicolas and Maeder,  Alexander and Maillou,  Vincent and Winka,  Anders and Cao,  Jiang and Kwasniewski,  Grzegorz and Deuschle,  Leonard and Hoefler,  Torsten and Ziogas,  Alexandros Nikolaos and Luisier,  Mathieu},
  year = {2025},
  month = nov,
  pages = {1–13},
  collection = {SC ’25}
}

@misc{omol,
  doi = {10.48550/ARXIV.2505.08762},
  url = {https://doi.org/10.48550/ARXIV.2505.08762},
  author = {Levine,  Daniel S. and Shuaibi,  Muhammed and Spotte-Smith,  Evan Walter Clark and Taylor,  Michael G. and Hasyim,  Muhammad R. and Michel,  Kyle and Batatia,  Ilyes and Csányi,  Gábor and Dzamba,  Misko and Eastman,  Peter and Frey,  Nathan C. and Fu,  Xiang and Gharakhanyan,  Vahe and Krishnapriyan,  Aditi S. and Rackers,  Joshua A. and Raja,  Sanjeev and Rizvi,  Ammar and Rosen,  Andrew S. and Ulissi,  Zachary and Vargas,  Santiago and Zitnick,  C. Lawrence and Blau,  Samuel M. and Wood,  Brandon M.},
  title = {The Open Molecules 2025 (OMol25) Dataset,  Evaluations,  and Models},
  publisher = {arXiv},
  year = {2025},
  copyright = {arXiv.org perpetual,  non-exclusive license}
}

@article{mace,
    author = {Batatia, Ilyes and Benner, Philipp and Chiang, Yuan and Elena, Alin M. and Kovács, Dávid P. and Riebesell, Janosh and Advincula, Xavier R. and Asta, Mark and Avaylon, Matthew and Baldwin, William J. and Berger, Fabian and Bernstein, Noam and Bhowmik, Arghya and Bigi, Filippo and Blau, Samuel M. and Cărare, Vlad and Ceriotti, Michele and Chong, Sanggyu and Darby, James P. and De, Sandip and Della Pia, Flaviano and Deringer, Volker L. and Elijošius, Rokas and El-Machachi, Zakariya and Fako, Edvin and Falcioni, Fabio and Ferrari, Andrea C. and Gardner, John L. A. and Gawkowski, Mikołaj J. and Genreith-Schriever, Annalena and George, Janine and Goodall, Rhys E. A. and Grandel, Jonas and Grey, Clare P. and Grigorev, Petr and Han, Shuang and Handley, Will and Heenen, Hendrik H. and Hermansson, Kersti and Ho, Cheuk Hin and Hofmann, Stephan and Holm, Christian and Jaafar, Jad and Jakob, Konstantin S. and Jung, Hyunwook and Kapil, Venkat and Kaplan, Aaron D. and Karimitari, Nima and Kermode, James R. and Kourtis, Panagiotis and Kroupa, Namu and Kullgren, Jolla and Kuner, Matthew C. and Kuryla, Domantas and Liepuoniute, Guoda and Lin, Chen and Margraf, Johannes T. and Magdău, Ioan-Bogdan and Michaelides, Angelos and Moore, J. Harry and Naik, Aakash A. and Niblett, Samuel P. and Norwood, Sam Walton and O’Neill, Niamh and Ortner, Christoph and Persson, Kristin A. and Reuter, Karsten and Rosen, Andrew S. and Rosset, Louise A. M. and Schaaf, Lars L. and Schran, Christoph and Shi, Benjamin X. and Sivonxay, Eric and Stenczel, Tamás K. and Sutton, Christopher and Svahn, Viktor and Swinburne, Thomas D. and Tilly, Jules and van der Oord, Cas and Vargas, Santiago and Varga-Umbrich, Eszter and Vegge, Tejs and Vondrák, Martin and Wang, Yangshuai and Witt, William C. and Wolf, Thomas and Zills, Fabian and Csányi, Gábor},
    title = {A foundation model for atomistic materials chemistry},
    journal = {The Journal of Chemical Physics},
    volume = {163},
    number = {18},
    pages = {184110},
    year = {2025},
    month = {11},
    issn = {0021-9606},
    doi = {10.1063/5.0297006},
    url = {https://doi.org/10.1063/5.0297006},
    eprint = {https://pubs.aip.org/aip/jcp/article-pdf/doi/10.1063/5.0297006/20801246/184110_1_5.0297006.pdf},
}

@inproceedings{uma,
title={{UMA}: A Family of Universal Models for Atoms},
author={Brandon M Wood and Misko Dzamba and Xiang Fu and Meng Gao and Muhammed Shuaibi and Luis Barroso-Luque and Kareem Abdelmaqsoud and Vahe Gharakhanyan and John R. Kitchin and Daniel S. Levine and Kyle Michel and Anuroop Sriram and Taco Cohen and Abhishek Das and Sushree Jagriti Sahoo and Ammar Rizvi and Zachary Ward Ulissi and C. Lawrence Zitnick},
booktitle={The Thirty-ninth Annual Conference on Neural Information Processing Systems},
year={2025},
url={https://openreview.net/forum?id=SvopaNxYWt}
}

@inproceedings{wanet,
title={Enhancing the Scalability and Applicability of Kohn-Sham Hamiltonians for Molecular Systems},
author={Yunyang Li and Zaishuo Xia and Lin Huang and Xinran Wei and Samuel Harshe and Han Yang and Erpai Luo and Zun Wang and Jia Zhang and Chang Liu and Bin Shao and Mark Gerstein},
booktitle={The Thirteenth International Conference on Learning Representations},
year={2025},
url={https://openreview.net/forum?id=twEvvkQqPS}
}

@inproceedings{helm,
title={Learning from the Electronic Structure of Molecules across the Periodic Table},
author={Manasa Kaniselvan and Benjamin Kurt Miller and Meng Gao and Juno Nam and Daniel S. Levine},
booktitle={The Fourteenth International Conference on Learning Representations},
year={2026},
url={https://openreview.net/forum?id=PS1YS8Wv4t}
}

@article{Suman2025,
  title = {Exploring the Design Space of Machine Learning Models for Quantum Chemistry with a Fully Differentiable Framework},
  volume = {21},
  ISSN = {1549-9626},
  url = {http://dx.doi.org/10.1021/acs.jctc.5c00522},
  DOI = {10.1021/acs.jctc.5c00522},
  number = {13},
  journal = {Journal of Chemical Theory and Computation},
  publisher = {American Chemical Society (ACS)},
  author = {Suman,  Divya and Nigam,  Jigyasa and Saade,  Sandra and Pegolo,  Paolo and T\"{u}rk,  Hanna and Zhang,  Xing and Chan,  Garnet Kin-Lic and Ceriotti,  Michele},
  year = {2025},
  month = jun,
  pages = {6505–6516}
}

@misc{hip,
  doi = {10.48550/ARXIV.2509.21624},
  url = {https://doi.org/10.48550/ARXIV.2509.21624},
  author = {Burger,  Andreas and Thiede,  Luca and Rønne,  Nikolaj and Bernales,  Varinia and Vijaykumar,  Nandita and Vegge,  Tejs and Bhowmik,  Arghya and Aspuru-Guzik,  Alan},
  title = {Shoot from the HIP: Hessian Interatomic Potentials without derivatives},
  publisher = {arXiv},
  year = {2025},
  copyright = {Creative Commons Attribution Non Commercial Share Alike 4.0 International}
}

@misc{deeph2,
  doi = {10.48550/ARXIV.2401.17015},
  url = {https://doi.org/10.48550/ARXIV.2401.17015},
  author = {Wang,  Yuxiang and Li,  He and Tang,  Zechen and Tao,  Honggeng and Wang,  Yanzhen and Yuan,  Zilong and Chen,  Zezhou and Duan,  Wenhui and Xu,  Yong},
  title = {{DeepH-2}: Enhancing deep-learning electronic structure via an equivariant local-coordinate transformer},
  publisher = {arXiv},
  year = {2024},
  copyright = {arXiv.org perpetual,  non-exclusive license}
}

@inproceedings{qh9dataset,
 author = {Yu, Haiyang and Liu, Meng and Luo, Youzhi and Strasser, Alex and Qian, Xiaofeng and Qian, Xiaoning and Ji, Shuiwang},
 booktitle = {Advances in Neural Information Processing Systems},
 editor = {A. Oh and T. Naumann and A. Globerson and K. Saenko and M. Hardt and S. Levine},
 pages = {40487--40503},
 publisher = {Curran Associates, Inc.},
 title = {QH9: A Quantum Hamiltonian Prediction Benchmark for QM9 Molecules},
 url = {https://proceedings.neurips.cc/paper_files/paper/2023/file/7f755e271717450020fda40f020922dd-Paper-Datasets_and_Benchmarks.pdf},
 volume = {36},
 year = {2023}
}

@misc{e3nn,
  doi = {10.48550/ARXIV.2207.09453},
  url = {https://doi.org/10.48550/ARXIV.2207.09453},
  author = {Geiger,  Mario and Smidt,  Tess},
  title = {e3nn: Euclidean Neural Networks},
  publisher = {arXiv},
  year = {2022},
  copyright = {Creative Commons Attribution 4.0 International}
}

@article{Tang2024,
  title = {Approaching coupled-cluster accuracy for molecular electronic structures with multi-task learning},
  volume = {5},
  ISSN = {2662-8457},
  url = {http://dx.doi.org/10.1038/s43588-024-00747-9},
  DOI = {10.1038/s43588-024-00747-9},
  number = {2},
  journal = {Nature Computational Science},
  publisher = {Springer Science and Business Media LLC},
  author = {Tang,  Hao and Xiao,  Brian and He,  Wenhao and Subasic,  Pero and Harutyunyan,  Avetik R. and Wang,  Yao and Liu,  Fang and Xu,  Haowei and Li,  Ju},
  year = {2024},
  month = dec,
  pages = {144–154}
}

@article{Brandbyge2002,
  title = {Density-functional method for nonequilibrium electron transport},
  volume = {65},
  ISSN = {1095-3795},
  url = {http://dx.doi.org/10.1103/PhysRevB.65.165401},
  DOI = {10.1103/physrevb.65.165401},
  number = {16},
  journal = {Physical Review B},
  publisher = {American Physical Society (APS)},
  author = {Brandbyge,  Mads and Mozos,  José-Luis and Ordejón,  Pablo and Taylor,  Jeremy and Stokbro,  Kurt},
  year = {2002},
  month = mar 
}

@inproceedings{Yeap2024,
  title = {2nm Platform Technology Featuring Energy-Efficient Nanosheet Transistors and Interconnects Co-Optimized with {3DIC} for {AI},  {HPC} and Mobile {SoC} Applications},
  url = {http://dx.doi.org/10.1109/IEDM50854.2024.10873475},
  DOI = {10.1109/iedm50854.2024.10873475},
  booktitle = {2024 IEEE International Electron Devices Meeting (IEDM)},
  publisher = {IEEE},
  author = {Yeap,  Geoffrey and Lin,  S.S. and Shang,  H.L. and Lin,  H.C. and Peng,  Y.C. and Wang,  M. and Wang,  PW and Lin,  CP and Yu,  KF and Lee,  WY and Chen,  HK and Lin,  DW and Yang,  BR and Yeh,  CC and Chan,  CT and Kuo,  JM and Liu,  C-M and Chiu,  TH and Wen,  MC and Lee,  T.L. and Chang,  CY and Chen,  R. and Huang,  P-H and Hou,  C.S. and Lin,  YK and Yang,  FK and Wang,  J. and Fung,  S. and Chen,  Ryan and Lee,  C.H. and Lee,  TL and Chang,  W. and Lee,  DY and Ting,  CY and Chang,  T. and Huang,  HC and Lin,  HJ and Tseng,  C. and Chang,  CW and Huang,  KB and Lu,  YC and Chen,  C-H and Chui,  C.O. and Chen,  KW and Tsai,  MH and Chen,  CC and Wu,  N. and Chiang,  HT and Chen,  XM and Sun,  SH and Tzeng,  JT and Wang,  K. and Peng,  YC and Liao,  HJ and Chen,  T. and Cheng,  YK and Chang,  J. and Hsieh,  K. and Cheng,  A. and Liu,  G. and Chen,  A. and Lin,  HT and Chiang,  KC and Tsai,  CW and Wang,  H. and Sheu,  W. and Yeh,  J. and Chen,  YM and Lin,  CK and Wu,  J. and Cao,  M. and Juang,  LS and Lai,  F. and Ku,  Y. and Jang,  S.M. and Lu,  L.C.},
  year = {2024},
  month = dec,
  pages = {1–4}
}

@article{Park2024,
  title = {Scalable Parallel Algorithm for Graph Neural Network Interatomic Potentials in Molecular Dynamics Simulations},
  volume = {20},
  ISSN = {1549-9626},
  url = {http://dx.doi.org/10.1021/acs.jctc.4c00190},
  DOI = {10.1021/acs.jctc.4c00190},
  number = {11},
  journal = {Journal of Chemical Theory and Computation},
  publisher = {American Chemical Society (ACS)},
  author = {Park,  Yutack and Kim,  Jaesun and Hwang,  Seungwoo and Han,  Seungwu},
  year = {2024},
  month = may,
  pages = {4857–4868}
}

@misc{deeph-pack,
  doi = {10.48550/ARXIV.2601.02938},
  url = {https://doi.org/10.48550/ARXIV.2601.02938},
  author = {Li,  Yang and Wang,  Yanzhen and Zhao,  Boheng and Gong,  Xiaoxun and Wang,  Yuxiang and Tang,  Zechen and Wang,  Zixu and Yuan,  Zilong and Li,  Jialin and Sun,  Minghui and Chen,  Zezhou and Tao,  Honggeng and Wu,  Baochun and Yu,  Yuhang and Li,  He and da Jornada,  Felipe H. and Duan,  Wenhui and Xu,  Yong},
  title = {DeepH-pack: A general-purpose neural network package for deep-learning electronic structure calculations},
  publisher = {arXiv},
  year = {2026},
  copyright = {arXiv.org perpetual,  non-exclusive license}
}

@inproceedings{amorphous,
title={Learning the Electronic Hamiltonian of Large Atomic Structures},
author={Chen Hao Xia and Manasa Kaniselvan and Alexandros Nikolaos Ziogas and Marko Mladenovi{\'c} and Rayen Mahjoub and Alexander Maeder and Mathieu Luisier},
booktitle={Forty-second International Conference on Machine Learning},
year={2025},
url={https://openreview.net/forum?id=WGejWCgrpD}
}

@article{Batzner2022,
  title = {{E(3)}-equivariant graph neural networks for data-efficient and accurate interatomic potentials},
  volume = {13},
  ISSN = {2041-1723},
  url = {http://dx.doi.org/10.1038/s41467-022-29939-5},
  DOI = {10.1038/s41467-022-29939-5},
  number = {1},
  journal = {Nature Communications},
  publisher = {Springer Science and Business Media LLC},
  author = {Batzner,  Simon and Musaelian,  Albert and Sun,  Lixin and Geiger,  Mario and Mailoa,  Jonathan P. and Kornbluth,  Mordechai and Molinari,  Nicola and Smidt,  Tess E. and Kozinsky,  Boris},
  year = {2022},
  month = may 
}

@article{brehmer2024doesequivariancematterscale,
title={Does equivariance matter at scale?},
author={Johann Brehmer and S{\"o}nke Behrends and Pim De Haan and Taco Cohen},
journal={Transactions on Machine Learning Research},
issn={2835-8856},
year={2025},
url={https://openreview.net/forum?id=wilNute8Tn},
note={}
}

@article{Kohn1965,
  title = {Self-Consistent Equations Including Exchange and Correlation Effects},
  volume = {140},
  ISSN = {0031-899X},
  url = {http://dx.doi.org/10.1103/PhysRev.140.A1133},
  DOI = {10.1103/physrev.140.a1133},
  number = {4A},
  journal = {Physical Review},
  publisher = {American Physical Society (APS)},
  author = {Kohn,  W. and Sham,  L. J.},
  year = {1965},
  month = nov,
  pages = {A1133–A1138}
}

@inproceedings{deepmd,
author = {Jia, Weile and Wang, Han and Chen, Mohan and Lu, Denghui and Lin, Lin and Car, Roberto and E, Weinan and Zhang, Linfeng},
title = {Pushing the limit of molecular dynamics with ab initio accuracy to 100 million atoms with machine learning},
year = {2020},
isbn = {9781728199986},
publisher = {IEEE Press},
booktitle = {Proceedings of the International Conference for High Performance Computing, Networking, Storage and Analysis},
articleno = {5},
numpages = {14},
location = {Atlanta, Georgia},
series = {SC '20},
url={https://dl.acm.org/doi/abs/10.5555/3433701.3433707}
}

@inproceedings{Kozinsky2023,
  series = {SC ’23},
  title = {Scaling the Leading Accuracy of Deep Equivariant Models to Biomolecular Simulations of Realistic Size},
  url = {http://dx.doi.org/10.1145/3581784.3627041},
  DOI = {10.1145/3581784.3627041},
  booktitle = {Proceedings of the International Conference for High Performance Computing,  Networking,  Storage and Analysis},
  publisher = {ACM},
  author = {Kozinsky,  Boris and Musaelian,  Albert and Johansson,  Anders and Batzner,  Simon},
  year = {2023},
  month = nov,
  pages = {1–12},
  collection = {SC ’23}
}

@InProceedings{so2,
  title = 	 {Reducing {SO}(3) Convolutions to {SO}(2) for Efficient Equivariant {GNN}s},
  author =       {Passaro, Saro and Zitnick, C. Lawrence},
  booktitle = 	 {Proceedings of the 40th International Conference on Machine Learning},
  pages = 	 {27420--27438},
  year = 	 {2023},
  editor = 	 {Krause, Andreas and Brunskill, Emma and Cho, Kyunghyun and Engelhardt, Barbara and Sabato, Sivan and Scarlett, Jonathan},
  volume = 	 {202},
  series = 	 {Proceedings of Machine Learning Research},
  month = 	 {23--29 Jul},
  publisher =    {PMLR},
  url = 	 {https://proceedings.mlr.press/v202/passaro23a.html}
}

@Article{LAMMPS,
  author = "A. P. Thompson and H. M. Aktulga and R. Berger and 
     D. S. Bolintineanu and W. M. Brown and P. S. Crozier and
     P. J. in 't Veld and A. Kohlmeyer and S. G. Moore and T. D. Nguyen and
     R. Shan and M. J. Stevens and J. Tranchida and C. Trott and S. J. Plimpton",
  title = "{LAMMPS} - a Flexible Simulation Tool for
     Particle-Based Materials Modeling at the 
     Atomic, Meso, and Continuum Scales",
  journal = "Comput. Phys. Commun.",
  volume =  "271",
  pages =   "108171",
  year =    "2022",
  url = "https://doi.org/10.1016/j.cpc.2021.108171"
}

@inproceedings{distmlip,
title={Dist{MLIP}: A Distributed Inference Platform for Machine Learning Interatomic Potentials},
author={Kevin Han and Bowen Deng and Amir Barati Farimani and Gerbrand Ceder},
booktitle={The Fourteenth International Conference on Learning Representations},
year={2026},
url={https://openreview.net/forum?id=4tasfBIPxp}
}

@article{deeptb,
  title = {Deep learning tight-binding approach for large-scale electronic simulations at finite temperatures with ab initio accuracy},
  volume = {15},
  ISSN = {2041-1723},
  url = {http://dx.doi.org/10.1038/s41467-024-51006-4},
  DOI = {10.1038/s41467-024-51006-4},
  number = {1},
  journal = {Nature Communications},
  publisher = {Springer Science and Business Media LLC},
  author = {Gu,  Qiangqiang and Zhouyin,  Zhanghao and Pandey,  Shishir Kumar and Zhang,  Peng and Zhang,  Linfeng and E,  Weinan},
  year = {2024},
  month = aug 
}

@article{hamster,
  title = {Physics-informed Hamiltonian learning for large-scale optoelectronic property prediction},
  volume = {17},
  ISSN = {2041-1723},
  url = {http://dx.doi.org/10.1038/s41467-026-70865-7},
  DOI = {10.1038/s41467-026-70865-7},
  number = {1},
  journal = {Nature Communications},
  publisher = {Springer Science and Business Media LLC},
  author = {Schwade,  Martin and Zhang,  Shaoming and Vonhoff,  Frederik and Delgado,  Frederico P. and Egger,  David A.},
  year = {2026},
  month = mar 
}

@misc{nigam2026,
  doi = {10.48550/ARXIV.2602.15345},
  url = {https://doi.org/10.48550/ARXIV.2602.15345},
  author = {Nigam,  Jigyasa and Smidt,  Tess and Dusson,  Geneviève},
  title = {Machine learning electronic structure and atomistic properties from the external potential},
  publisher = {arXiv},
  year = {2026},
  copyright = {Creative Commons Attribution 4.0 International}
}

@article{Yuan2026,
  title = {Foundation models for atomistic simulation of chemistry and materials},
  volume = {10},
  ISSN = {2397-3358},
  url = {http://dx.doi.org/10.1038/s41570-025-00793-5},
  DOI = {10.1038/s41570-025-00793-5},
  number = {3},
  journal = {Nature Reviews Chemistry},
  publisher = {Springer Science and Business Media LLC},
  author = {Yuan,  Eric C.-Y. and Liu,  Yunsheng and Chen,  Junmin and Zhong,  Peichen and Raja,  Sanjeev and Kreiman,  Tobias and Vargas,  Santiago and Xu,  Wenbin and Head-Gordon,  Martin and Yang,  Chao and Blau,  Samuel M. and Cheng,  Bingqing and Krishnapriyan,  Aditi and Head-Gordon,  Teresa},
  year = {2026},
  month = feb,
  pages = {212–230}
}

@misc{better_cuet,
  doi = {10.48550/ARXIV.2501.13986},
  url = {https://doi.org/10.48550/ARXIV.2501.13986},
  author = {Bharadwaj,  Vivek and Glover,  Austin and Buluc,  Aydin and Demmel,  James},
  title = {An Efficient Sparse Kernel Generator for O(3)-Equivariant Deep Networks},
  publisher = {arXiv},
  year = {2025},
  copyright = {Creative Commons Attribution 4.0 International}
}

@inproceedings{priceoffreedom,
title={The Price of Freedom: Exploring Expressivity and Runtime Tradeoffs in Equivariant Tensor Products},
author={YuQing Xie and Ameya Daigavane and Mit Kotak and Tess Smidt},
booktitle={Forty-second International Conference on Machine Learning},
year={2025},
url={https://openreview.net/forum?id=EvIwwGYTLc}
}

@article{pet,
  title = {PET-MAD as a lightweight universal interatomic potential for advanced materials modeling},
  volume = {16},
  ISSN = {2041-1723},
  url = {http://dx.doi.org/10.1038/s41467-025-65662-7},
  DOI = {10.1038/s41467-025-65662-7},
  number = {1},
  journal = {Nature Communications},
  publisher = {Springer Science and Business Media LLC},
  author = {Mazitov,  Arslan and Bigi,  Filippo and Kellner,  Matthias and Pegolo,  Paolo and Tisi,  Davide and Fraux,  Guillaume and Pozdnyakov,  Sergey and Loche,  Philip and Ceriotti,  Michele},
  year = {2025},
  month = nov 
}

@article{Christensen2020,
author = "Anders S. Christensen and Anatole Von lilienfeld",
title = "{Revised MD17 dataset (rMD17)}",
year = "2020",
month = "7",
doi = "10.6084/m9.figshare.12672038.v3",
url = "https://doi.org/10.6084/m9.figshare.12672038.v3"
}

@article{deeph3,
  title = {General framework for {E(3)}-equivariant neural network representation of density functional theory Hamiltonian},
  volume = {14},
  ISSN = {2041-1723},
  url = {http://dx.doi.org/10.1038/s41467-023-38468-8},
  DOI = {10.1038/s41467-023-38468-8},
  number = {1},
  journal = {Nature Communications},
  publisher = {Springer Science and Business Media LLC},
  author = {Gong,  Xiaoxun and Li,  He and Zou,  Nianlong and Xu,  Runzhang and Duan,  Wenhui and Xu,  Yong},
  year = {2023},
  month = may 
}

@article{Kaniselvan2023,
  title = {An Atomistic Model of Field-Induced Resistive Switching in Valence Change Memory},
  volume = {17},
  ISSN = {1936-086X},
  url = {http://dx.doi.org/10.1021/acsnano.2c12575},
  DOI = {10.1021/acsnano.2c12575},
  number = {9},
  journal = {ACS Nano},
  publisher = {American Chemical Society (ACS)},
  author = {Kaniselvan,  Manasa and Luisier,  Mathieu and Mladenović,  Marko},
  year = {2023},
  month = mar,
  pages = {8281–8292}
}

@InProceedings{eSEN,
  title = 	 {Learning Smooth and Expressive Interatomic Potentials for Physical Property Prediction},
  author =       {Fu, Xiang and Wood, Brandon M and Barroso-Luque, Luis and Levine, Daniel S. and Gao, Meng and Dzamba, Misko and Zitnick, C. Lawrence},
  booktitle = 	 {Proceedings of the 42nd International Conference on Machine Learning},
  pages = 	 {17875--17893},
  year = 	 {2025},
  editor = 	 {Singh, Aarti and Fazel, Maryam and Hsu, Daniel and Lacoste-Julien, Simon and Berkenkamp, Felix and Maharaj, Tegan and Wagstaff, Kiri and Zhu, Jerry},
  volume = 	 {267},
  series = 	 {Proceedings of Machine Learning Research},
  month = 	 {13--19 Jul},
  publisher =    {PMLR},
  url = 	 {https://proceedings.mlr.press/v267/fu25h.html}
}

@article{metis,
  title = {Multilevelk-way Partitioning Scheme for Irregular Graphs},
  volume = {48},
  ISSN = {0743-7315},
  url = {http://dx.doi.org/10.1006/jpdc.1997.1404},
  DOI = {10.1006/jpdc.1997.1404},
  number = {1},
  journal = {Journal of Parallel and Distributed Computing},
  publisher = {Elsevier BV},
  author = {Karypis,  George and Kumar,  Vipin},
  year = {1998},
  month = jan,
  pages = {96–129}
}

@article{schnorb,
  title = {Unifying machine learning and quantum chemistry with a deep neural network for molecular wavefunctions},
  volume = {10},
  ISSN = {2041-1723},
  url = {http://dx.doi.org/10.1038/s41467-019-12875-2},
  DOI = {10.1038/s41467-019-12875-2},
  number = {1},
  journal = {Nature Communications},
  publisher = {Springer Science and Business Media LLC},
  author = {Sch\"{u}tt,  K. T. and Gastegger,  M. and Tkatchenko,  A. and M\"{u}ller,  K.-R. and Maurer,  R. J.},
  year = {2019},
  month = nov 
}

@article{Dossena2025,
  title = {Mobility calculation in disordered WS2-Al2O3 stacks from first principles},
  volume = {9},
  ISSN = {2397-7132},
  url = {http://dx.doi.org/10.1038/s41699-025-00587-9},
  DOI = {10.1038/s41699-025-00587-9},
  number = {1},
  journal = {npj 2D Materials and Applications},
  publisher = {Springer Science and Business Media LLC},
  author = {Dossena,  Mauro and Van Troeye,  Benoit and Ducry,  Fabian and Cao,  Jiang and Afzalian,  Aryan and Pourtois,  Geoffrey and Luisier,  Mathieu},
  year = {2025},
  month = aug 
}

@article{Onofrio2015,
  title = {Atomic origin of ultrafast resistance switching in nanoscale electrometallization cells},
  volume = {14},
  ISSN = {1476-4660},
  url = {http://dx.doi.org/10.1038/nmat4221},
  DOI = {10.1038/nmat4221},
  number = {4},
  journal = {Nature Materials},
  publisher = {Springer Science and Business Media LLC},
  author = {Onofrio,  Nicolas and Guzman,  David and Strachan,  Alejandro},
  year = {2015},
  month = mar,
  pages = {440–446}
}

@article{Zhou2023,
  title = {Device-scale atomistic modelling of phase-change memory materials},
  volume = {6},
  ISSN = {2520-1131},
  url = {http://dx.doi.org/10.1038/s41928-023-01030-x},
  DOI = {10.1038/s41928-023-01030-x},
  number = {10},
  journal = {Nature Electronics},
  publisher = {Springer Science and Business Media LLC},
  author = {Zhou,  Yuxing and Zhang,  Wei and Ma,  En and Deringer,  Volker L.},
  year = {2023},
  month = sep,
  pages = {746–754}
}

@article{Nigam2022,
  title = {Equivariant representations for molecular Hamiltonians and N-center atomic-scale properties},
  volume = {156},
  ISSN = {1089-7690},
  url = {http://dx.doi.org/10.1063/5.0072784},
  DOI = {10.1063/5.0072784},
  number = {1},
  journal = {The Journal of Chemical Physics},
  publisher = {AIP Publishing},
  author = {Nigam,  Jigyasa and Willatt,  Michael J. and Ceriotti,  Michele},
  year = {2022},
  month = jan 
}

@InProceedings{quantumham,
  title = 	 {Efficient and Equivariant Graph Networks for Predicting Quantum {H}amiltonian},
  author =       {Yu, Haiyang and Xu, Zhao and Qian, Xiaofeng and Qian, Xiaoning and Ji, Shuiwang},
  booktitle = 	 {Proceedings of the 40th International Conference on Machine Learning},
  pages = 	 {40412--40424},
  year = 	 {2023},
  editor = 	 {Krause, Andreas and Brunskill, Emma and Cho, Kyunghyun and Engelhardt, Barbara and Sabato, Sivan and Scarlett, Jonathan},
  volume = 	 {202},
  series = 	 {Proceedings of Machine Learning Research},
  month = 	 {23--29 Jul},
  publisher =    {PMLR},
  url = 	 {https://proceedings.mlr.press/v202/yu23i.html}
}

@article{deeph,
  title = {Deep-learning density functional theory Hamiltonian for efficient ab initio electronic-structure calculation},
  volume = {2},
  ISSN = {2662-8457},
  url = {http://dx.doi.org/10.1038/s43588-022-00265-6},
  DOI = {10.1038/s43588-022-00265-6},
  number = {6},
  journal = {Nature Computational Science},
  publisher = {Springer Science and Business Media LLC},
  author = {Li,  He and Wang,  Zun and Zou,  Nianlong and Ye,  Meng and Xu,  Runzhang and Gong,  Xiaoxun and Duan,  Wenhui and Xu,  Yong},
  year = {2022},
  month = jun,
  pages = {367–377}
}

@article{orca,
  title = {The ORCA program system},
  volume = {2},
  ISSN = {1759-0884},
  url = {http://dx.doi.org/10.1002/wcms.81},
  DOI = {10.1002/wcms.81},
  number = {1},
  journal = {WIREs Computational Molecular Science},
  publisher = {Wiley},
  author = {Neese,  Frank},
  year = {2011},
  month = jun,
  pages = {73–78}
}

@misc{hoefler_alps_bench,
  doi = {10.48550/ARXIV.2408.11556},
  url = {https://doi.org/10.48550/ARXIV.2408.11556},
  author = {Fusco,  Luigi and Khalilov,  Mikhail and Chrapek,  Marcin and Chukkapalli,  Giridhar and Schulthess,  Thomas and Hoefler,  Torsten},
  title = {Understanding Data Movement in Tightly Coupled Heterogeneous Systems: A Case Study with the Grace Hopper Superchip},
  publisher = {arXiv},
  year = {2024},
  copyright = {Creative Commons Attribution Share Alike 4.0 International}
}

@inproceedings{pytorch2024,
author = {Ansel, Jason and Yang, Edward and He, Horace and Gimelshein, Natalia and Jain, Animesh and Voznesensky, Michael and Bao, Bin and Bell, Peter and Berard, David and Burovski, Evgeni and Chauhan, Geeta and Chourdia, Anjali and Constable, Will and Desmaison, Alban and DeVito, Zachary and Ellison, Elias and Feng, Will and Gong, Jiong and Gschwind, Michael and Hirsh, Brian and Huang, Sherlock and Kalambarkar, Kshiteej and Kirsch, Laurent and Lazos, Michael and Lezcano, Mario and Liang, Yanbo and Liang, Jason and Lu, Yinghai and Luk, C. K. and Maher, Bert and Pan, Yunjie and Puhrsch, Christian and Reso, Matthias and Saroufim, Mark and Siraichi, Marcos Yukio and Suk, Helen and Zhang, Shunting and Suo, Michael and Tillet, Phil and Zhao, Xu and Wang, Eikan and Zhou, Keren and Zou, Richard and Wang, Xiaodong and Mathews, Ajit and Wen, William and Chanan, Gregory and Wu, Peng and Chintala, Soumith},
title = {PyTorch 2: Faster Machine Learning Through Dynamic Python Bytecode Transformation and Graph Compilation},
year = {2024},
isbn = {9798400703850},
publisher = {Association for Computing Machinery},
address = {New York, NY, USA},
url = {https://doi.org/10.1145/3620665.3640366},
doi = {10.1145/3620665.3640366},
booktitle = {Proceedings of the 29th ACM International Conference on Architectural Support for Programming Languages and Operating Systems, Volume 2},
pages = {929–947},
numpages = {19},
location = {La Jolla, CA, USA},
series = {ASPLOS '24}
}

@article{cp2k,
    author = {Kühne, Thomas D. and Iannuzzi, Marcella and Del Ben, Mauro and Rybkin, Vladimir V. and Seewald, Patrick and Stein, Frederick and Laino, Teodoro and Khaliullin, Rustam Z. and Schütt, Ole and Schiffmann, Florian and Golze, Dorothea and Wilhelm, Jan and Chulkov, Sergey and Bani-Hashemian, Mohammad Hossein and Weber, Valéry and Borštnik, Urban and Taillefumier, Mathieu and Jakobovits, Alice Shoshana and Lazzaro, Alfio and Pabst, Hans and Müller, Tiziano and Schade, Robert and Guidon, Manuel and Andermatt, Samuel and Holmberg, Nico and Schenter, Gregory K. and Hehn, Anna and Bussy, Augustin and Belleflamme, Fabian and Tabacchi, Gloria and Glöß, Andreas and Lass, Michael and Bethune, Iain and Mundy, Christopher J. and Plessl, Christian and Watkins, Matt and VandeVondele, Joost and Krack, Matthias and Hutter, Jürg},
    title = {CP2K: An electronic structure and molecular dynamics software package - Quickstep: Efficient and accurate electronic structure calculations},
    journal = {The Journal of Chemical Physics},
    volume = {152},
    number = {19},
    pages = {194103},
    year = {2020},
    month = {05},
    issn = {0021-9606},
    doi = {10.1063/5.0007045},
    url = {https://doi.org/10.1063/5.0007045},
    eprint = {https://pubs.aip.org/aip/jcp/article-pdf/doi/10.1063/5.0007045/16718133/194103\_1\_online.pdf}
}

@INPROCEEDINGS{intel-nrfet,
  author={Agrawal, A. and Chakraborty, W. and Li, W. and Ryu, H. and Markman, B. and Hoon, S. H. and Paul, R. K and Huang, C. Y. and Choi, S. M. and Rho, K. and Shu, A. and Iglesias, R. and Wallace, P. and Ghosh, S. and Cheong, K. L. and Hockel, J. L. and Thorman, R. and Baumgartel, L. and Shoer, L. and Mishra, V. and Berrada, S. and Ashita, A. and Weber, C. and Obradovic, B. and Oni, A. A. and Brooks, Z. and Franco, N. and Kavalieros, J. and Dewey, G.},
  booktitle={2024 IEEE International Electron Devices Meeting (IEDM)}, 
  title={Silicon RibbonFET CMOS at 6nm Gate Length}, 
  year={2024},
  url={https://doi.org/10.1109/IEDM50854.2024.10873367}}


\end{document}